\documentclass[letterpaper]{article}
\usepackage{aaai2026}
\nocopyright
\usepackage{times}
\usepackage{helvet}
\usepackage{courier}
\usepackage[hyphens]{url}
\usepackage{graphicx}
\usepackage[numbers,sort&compress]{natbib}
\usepackage{caption}
\usepackage{booktabs}
\usepackage{multirow}
\usepackage{amsmath}
\usepackage{amssymb}
\usepackage{subcaption}
\usepackage{enumitem}
\usepackage{longtable}
\usepackage[table]{xcolor}
\usepackage{pifont}
\usepackage{array}
\usepackage{tabularx}
\usepackage{listings}
\newcommand{\na}{--}
\newcommand{\cmark}{\textcolor{teal!60!black}{\ding{51}}}
\newcommand{\xmark}{\textcolor{red!75!black}{\ding{55}}}
\newcommand{\tabhead}[2]{\shortstack{\citep{#2}\\\textbf{#1}}}
\title{From Simple QA to Deep Research: \\
A Verifiable Benchmark Constructed through Iterative Task Evolution}
\author{
\textbf{Can Wang\textsuperscript{\rm 1,2}\thanks{Equal contribution.},
Haoran Chen\textsuperscript{\rm 2}\footnotemark[1],
Haowen Gao\textsuperscript{\rm 2,3},
Hao Ding\textsuperscript{\rm 4},
Zhaoyang Liu\textsuperscript{\rm 2}\thanks{Corresponding author.},
Zhiying Tu\textsuperscript{\rm 1}\footnotemark[2]}
}
\affiliations{
\textsuperscript{\rm 1}Shandong Key Laboratory of Digital Service Computing Technology and Systems \\
\textsuperscript{\rm 2}Alibaba Token Hub \\
\textsuperscript{\rm 3}State Key Laboratory of AI Safety, Institute of Computing Technology, Chinese Academy of Sciences \\
\textsuperscript{\rm 4}Department of Electrical and Electronic Engineering, The Hong Kong Polytechnic University
\vspace{0.18in}
}

\begin{document}

\maketitle

\begin{abstract}
Deep research benchmarks require expert-level tasks and reliable evaluation grounded in task-specific knowledge.
Existing benchmarks rely heavily on expert authoring or pre-existing human-authored materials, while fully automatic construction struggles to ensure consistent and traceable verification.
To address this gap, we introduce a verifiable benchmark of 500 deep research tasks spanning 31 topics and 10 major categories, with three query forms designed to probe complementary capabilities required for deep research.
The benchmark is constructed automatically using an iterative Explorer--Formalizer--Challenger pipeline that progressively transforms simple questions into deep research tasks. 
Concretely, each task is represented as a directed acyclic graph (DAG) of atomic steps and associated checkpoints, enabling the query, DAG, and rubrics to evolve together in a controlled manner.
Experiments demonstrate that the benchmark clearly discriminates among models and across query types, while its fact-grounded pointwise rubrics enable fine-grained, human-aligned, and stable evaluation. Our data, implementation, and results are publicly available at \url{https://github.com/chr6192/TaskEvolving.git}.
\end{abstract}

\section{Introduction}
Deep research has attracted growing attention as AI agents evolve from fluent text generators into smarter tool-augmented assistants.
Deep research allows multiple valid solving paths and responses, while producing high-quality responses requires domain knowledge. This openness and demand for expertise make fixed uniform evaluation difficult, calling instead for fine-grained evaluation with rubrics grounded in task-specific knowledge~\citep{deepresearcheval,deer,deepresearchbench,drarena}.

\begin{figure}[!t]
\centering
\includegraphics[width=0.92\columnwidth]{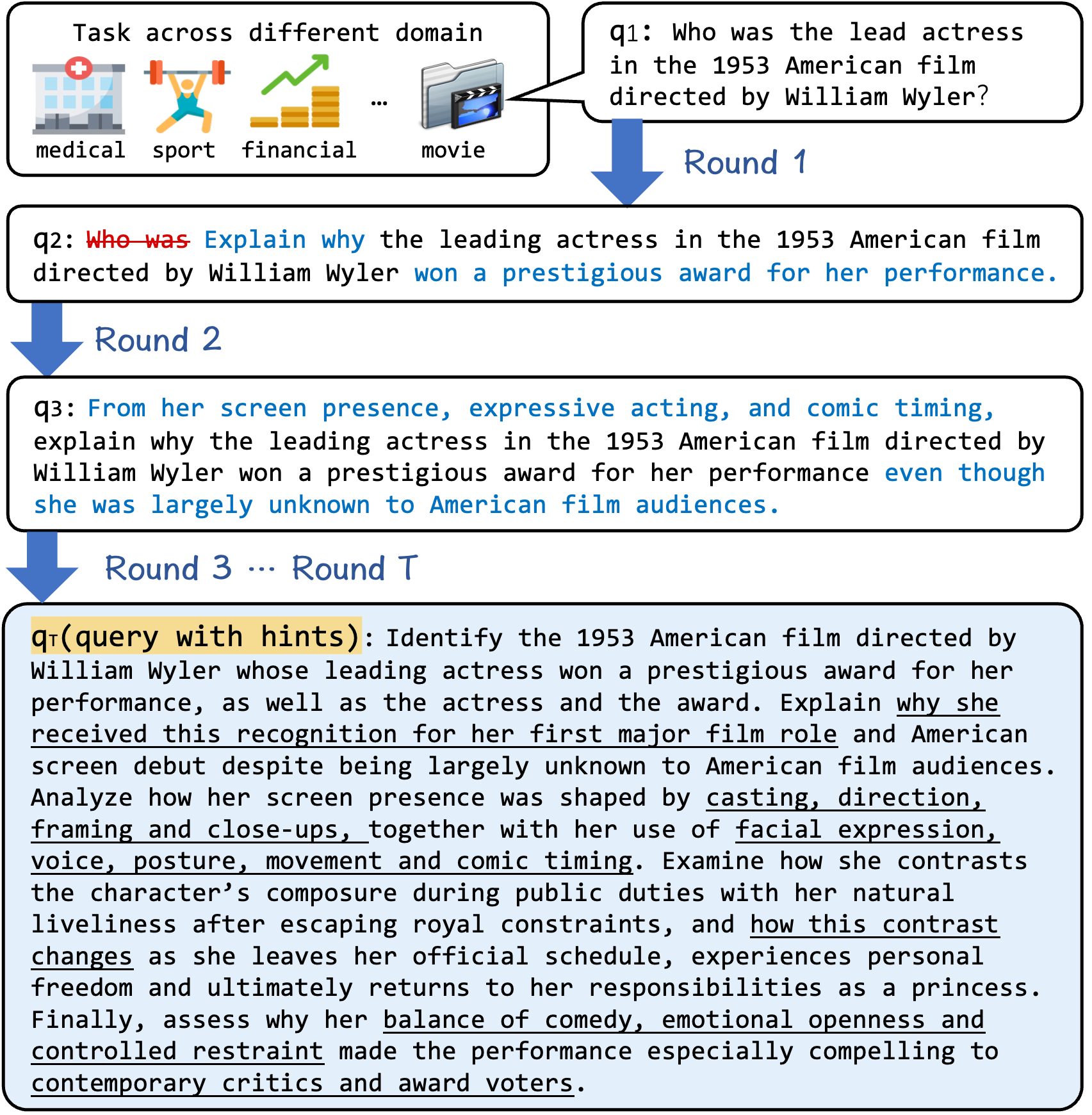}
\caption{Round~1 turns the simple query $q_1$ into a problem that requires one more step to solve.
Round~2 further evolves it into a query with multiple answering aspects.
Through further rounds up to Round~$T$, the process yields a substantially harder deep research task $q_T$ that requires multi-source search and integrative analysis.}
\label{fig:intro}
\end{figure}

However, automatically constructing deep research tasks with such evaluation is difficult. AgentDisCo, QUEST, and DR-Arena evaluate models by comparing their reports with one another~\citep{agentdisco,quest,drarenaframework}. This evaluation is useful for ranking models, but the resulting scores depend on which models are compared and do not provide an absolute measure of response quality. DeepResearchEval argues that generic evaluation dimensions are insufficient and uses an LLM to generate one to three task-specific rubrics for each query~\citep{deepresearcheval}. Yet the reliability and professional quality of the generated rubrics are not guaranteed.

\begin{table*}[!t]
\centering
\footnotesize
\setlength{\tabcolsep}{2.5pt}
\begin{tabularx}{\textwidth}{@{}>{\raggedright\arraybackslash}p{0.12\textwidth}%
>{\centering\arraybackslash}X%
>{\centering\arraybackslash}X%
>{\centering\arraybackslash}X%
>{\centering\arraybackslash}X%
>{\centering\arraybackslash}X%
>{\centering\arraybackslash}X%
>{\centering\arraybackslash}X%
>{\centering\arraybackslash}X%
>{\centering\arraybackslash}X%
>{\centering\arraybackslash}X%
@{\hspace{9pt}}%
>{\centering\arraybackslash}X%
>{\centering\arraybackslash}X%
>{\centering\arraybackslash}X%
@{\hspace{9pt}}%
>{\centering\arraybackslash}X%
>{\centering\arraybackslash}X%
>{\centering\arraybackslash}X@{}}
\toprule
\textbf{Dimension} & \tabhead{RR}{researchrubrics} & \tabhead{DEER}{deer} & \tabhead{FinR}{finresearchbench} & \tabhead{DR-I}{deepresearchbench} & \tabhead{MDR}{minddr} & \tabhead{Step}{stepdeepresearch} & \tabhead{DR-II}{drbench2} & \tabhead{RptB}{reportbench} & \tabhead{ADRA}{adrabank} & \tabhead{MMDR}{mmdrbench} & \tabhead{Arena}{drarena} & \tabhead{DisCo}{agentdisco} & \tabhead{QUEST}{quest} & \tabhead{DRA}{drarenaframework} & \tabhead{DREv}{deepresearcheval} & \textbf{Ours} \\
\midrule
Task structure & \xmark & \xmark & \cmark & \xmark & \cmark & \xmark & \xmark & \xmark & \cmark & \xmark & \xmark & \xmark & \xmark & \cmark & \xmark & \cmark \\
Difficulty ladder & \xmark & \xmark & \xmark & \xmark & \xmark & \xmark & \xmark & \xmark & \xmark & \xmark & \xmark & \cmark & \cmark & \xmark & \cmark & \cmark \\
Full automatic & \xmark & \xmark & \xmark & \xmark & \xmark & \xmark & \xmark & \xmark & \xmark & \xmark & \xmark & \cmark & \cmark & \cmark & \cmark & \cmark \\
Grounded rubric & \cmark & \cmark & \xmark & \xmark & \xmark & \xmark & \cmark & \cmark & \cmark & \xmark & \xmark & \xmark & \xmark & \cmark & \xmark & \cmark \\
Pointwise rubric & \cmark & \cmark & \cmark & \xmark & \xmark & \cmark & \cmark & \cmark & \cmark & \cmark & \cmark & \xmark & \xmark & \xmark & \cmark & \cmark \\
\bottomrule
\end{tabularx}
\caption{Comparison with representative deep research benchmarks on five dimensions.}
\label{tab:compare}
\end{table*}

We focus on the benchmark construction challenge: how to build a diverse collection of professionally deep research tasks together with reliable rubrics grounded in task-specific knowledge, without expert authoring or pre-written human materials.
Our key insight is that the task-specific knowledge needed for professional evaluation, including the evidence and reasoning a response should cover, can be progressively revealed as an agent expands its exploration scope.
Furthermore, solving a deep research task unfolds through continuous exploration in a partially observable environment, where an agent clarifies the goal, the scope of relevant evidence, and possible reasoning paths step by step~\cite{cgdp,agenticrlsurvey,webarena}. Task construction should therefore follow the same iterative process, rather than generating the query or rubrics in a single pass. By developing evaluation checkpoints alongside the expanding exploration scope, each checkpoint can be grounded in external evidence and enable reliable verification of task-specific claims.

To construct the benchmark, we use a task evolution pipeline that iteratively raises the difficulty of a simple query until it becomes a hard deep research task, as illustrated in Figure~\ref{fig:intro}.
In each round, an iterative workflow proceeds through three roles.
The \textbf{Explorer} solves the current task in an open environment and records the evidence and analysis it reaches.
The \textbf{Formalizer} organizes the task-relevant subset into a directed acyclic graph (DAG) that makes the solving process explicit as atomic research steps, and derives from each node an aligned rubric with verifiable checkpoints.
The \textbf{Challenger} identifies explored but unused clues that remain logically connected to the task and uses them to generate the next-round query.
Across rounds, this loop progressively incorporates newly found knowledge into the task, increasing difficulty without sacrificing coherence or solvability. As a result, the task, DAG, and rubrics evolve together in a coordinated manner. 

Our main contributions are threefold.
\begin{itemize}
\item We construct a benchmark of 500 professionally deep research tasks spanning 31 topics, each paired with fact-grounded pointwise rubrics and three query forms that probe complementary deep research abilities.
\item We provide a fully automatic construction pipeline centered on iterative DAG expansion, enabling the query, task structure, and rubrics to evolve together without relying on manual authoring or existing materials.
\item Extensive experiments demonstrate that the benchmark effectively reveals models’ strengths and weaknesses in deep research, with its rubrics enabling fine-grained, human-aligned, and stable evaluation.
\end{itemize}

\section{Related Work}

\subsection{From Deep Search to Deep Research}

Deep search aims to locate hard-to-find facts in noisy web environments through multi-round search, browsing, and tool use.
Representative deep search benchmarks typically construct hard short-answer tasks via human curation \citep{widesearch,mind2web2,mmina,gaia,browsecomp}, question linking \citep{webwalker,webdancer,deepresearch9k,toolhop}, or entity linking \citep{websailor,webshaper}.
They remain fact-seeking at core, where success means hitting independently verifiable short facts rather than integrating them into an open structured result.

Deep research addresses open-ended task needs that go beyond typical deep search tasks\citep{deepresearchsurvey,livedrbench}. 
First, deep search mainly stresses multi-hop retrieval, so difficulty can rise by chaining subquestions through question or entity links.
Deep research instead requires breaking down a research goal and integrating evidence into a coherent argument, so its subquestions must serve that goal and stay logically connected rather than form a loosely linked chain.
Second, a deep search task mainly anchors a truth source and collects the evidence needed to reach it.
Deep research builds on this and must further constrain the task scope: too broad and the research target becomes unbounded, too narrow and the task collapses into a checklist.
Third, deep search is evaluated by whether one unique fact is found.
Deep research remains open-ended, so neither valid solving paths nor qualified answers are unique, and evaluation must check whether an answer is correct without collapsing openness into matching one gold answer or procedure.

\begin{figure*}[!t]
\centering
\includegraphics[width=\textwidth]{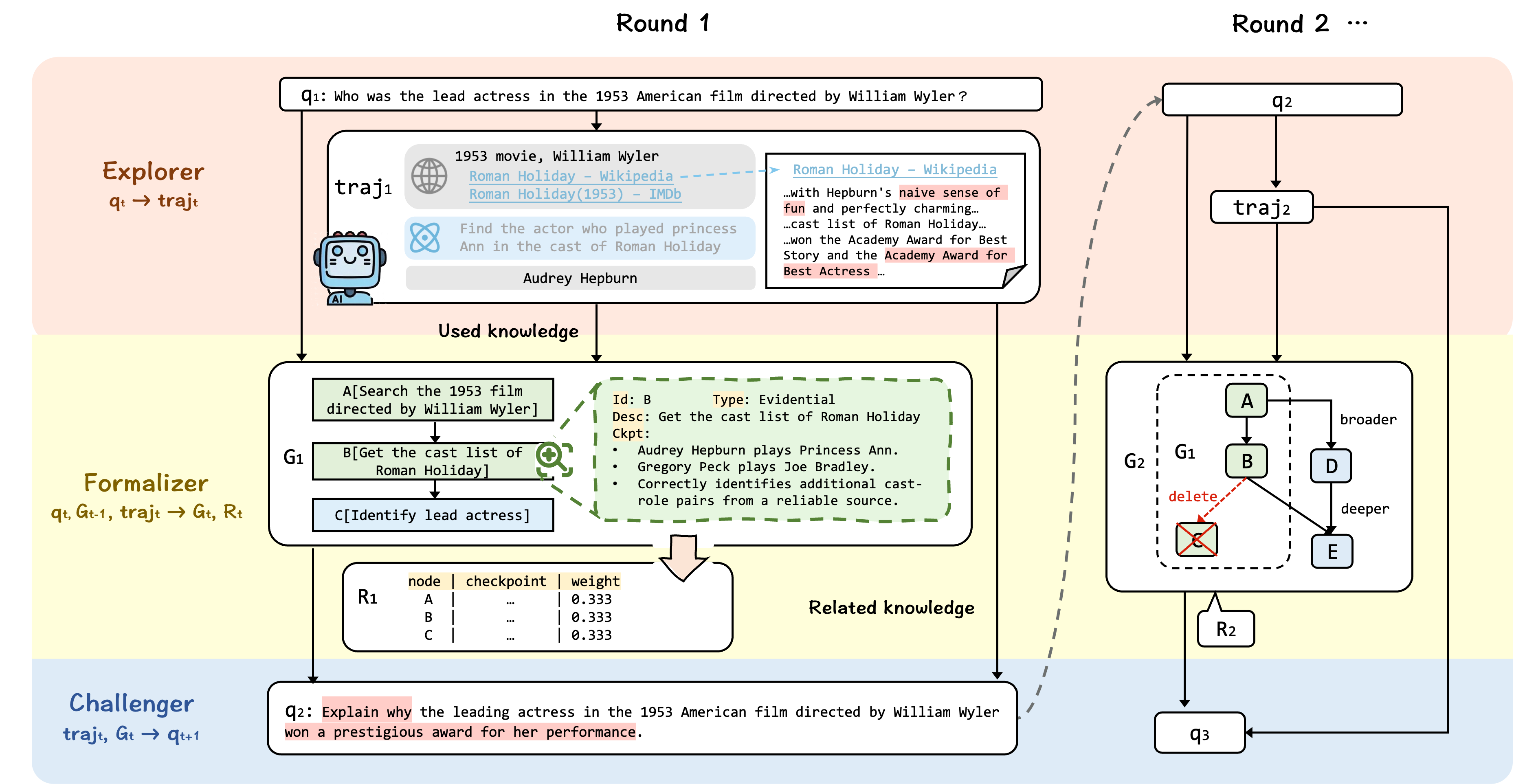}
\caption{Overview of our benchmark construction pipeline.
The Explorer, Formalizer, and Challenger loop evolves a simple query into a harder deep research task across rounds indexed by $t$. 
The figure shows one concrete step from $q_1$ to a harder $q_2$, with the task graph and rubrics updated accordingly, then continued expansion in round $t{=}2$.}
\label{fig:main}
\vspace{-2mm}
\end{figure*}

\subsection{Deep Research Benchmarks}

Deep research benchmarks span manual, semi-automatic, and automatic construction paradigms.
Manual benchmarks, represented by ResearchRubrics \citep{researchrubrics}, DEER \citep{deer}, and FinResearchBench \citep{finresearchbench}, rely on experts to write both queries and rubrics from scratch.
Semi-automatic benchmarks, represented by DR Bench~I \citep{deepresearchbench}, MindDR Bench \citep{minddr}, and Step-DeepResearch \citep{stepdeepresearch}, retain human-written or curated queries while assembling evaluation standards through automated pipelines.
Other semi-automatic benchmarks reverse-engineer task queries and evaluation rubrics from existing expert or human materials, including DR Bench~II \citep{drbench2}, ReportBench \citep{reportbench}, ADRA-Bank \citep{adrabank}, MMDR Bench \citep{mmdrbench}, and DeepResearch Arena \citep{drarena}.
These manual and semi-automatic benchmarks produce professional, reliable tasks but require substantial human effort or ready made expert materials, making them costly to extend to new domains.

Fully automatic benchmarks include AgentDisCo \citep{agentdisco}, QUEST \citep{quest}, DR-Arena \citep{drarenaframework}, and DeepResearchEval \citep{deepresearcheval}, which construct tasks without human participation.
AgentDisCo and QUEST derive tasks from user browsing behavior or trending queries, while DR-Arena builds dynamic tasks and grounded checklists from live web information.
All three evaluate agents through relative comparisons between reports.
Such comparisons support model ranking, but their scores depend on the models being compared and do not provide an absolute measure of individual response quality.
DeepResearchEval instead argues that generic deep research dimensions are insufficient and uses an LLM to generate one to three task-specific dimensions and rubrics per question.
Its experiments show that task-specific knowledge is a real weakness of current agents, but the reliability and professional quality of the generated rubrics remain difficult to guarantee.
Together, these benchmarks improve scalability and transferability while leaving reliable pointwise evaluation grounded in task-specific knowledge unresolved.

Table~\ref{tab:compare} compares benchmarks on five dimensions.
Compared with manual and semi-automatic benchmarks, ours requires neither expert-authored queries and rubrics nor ready made expert corpora.
Compared with automatic benchmarks, ours constructs tasks and rubrics through iterative interaction between an agent and its environment rather than generating them from prompts at once.
This keeps the resulting tasks and rubrics traceable to external evidence while allowing task difficulty to deepen across rounds.

\section{Benchmark Construction}

This section describes the pipeline used to construct our benchmark.
We first present the task representation, explaining why the DAG serves as the core structural representation of a deep research task.
We then introduce the Explorer, Formalizer, and Challenger loop, including the stopping criterion, and explain how the task and rubric evolve together across rounds.
Finally, we describe query design to test different deep research abilities of agents.

\subsection{Task Representation}

Our benchmark construction starts from a simple query $q_1$ and evolves it over $T$ rounds into a deep research task $q_T$ with domain knowledge, together with an aligned verifiable rubric set $R_T$ for evaluation.
The solving process of a deep research task follows a stable pattern of collecting evidence, integrating reasoning, and forming conclusions.
We therefore use a DAG $G=(V,E)$ as the structural representation and the core structure for evolution, which makes the solving process explicit, decomposable, and traceable.

Each node $v \in V$ represents an atomic research step and is written as $v = (id_v, desc_v, type_v, ckpt_v)$.
Here $id_v$ is a unique identifier.
$desc_v$ describes what the step does, such as collecting evidence or performing analysis.
$\mathrm{type}_v \in \{\mathcal{E}, \mathcal{A}\}$ labels the node as evidential ($\mathcal{E}$) or analytical ($\mathcal{A}$).
When $\mathrm{type}_v = \mathcal{E}$, the step retrieves traceable factual evidence from external sources, such as a policy release date or a company's revenue.
When $\mathrm{type}_v = \mathcal{A}$, the step performs analysis over already collected evidence, such as causal inference or cross-source comparison.
$ckpt_v = \{ckpt_{v1}, ckpt_{v2}, \ldots\}$ lists the checkpoints produced after executing node $v$, each of which can be independently verified, such as information retrieved by an evidential node or results derived by an analytical node.
Each edge $e=(u,v) \in E$ means that step $v$ depends on step $u$ logically and therefore cannot proceed until $u$ finishes.
Nodes that have no edge between them are parallelizable.
They typically cover different aspects of the same subtask and are linked only indirectly through shared neighbors, either branching from a common upstream node or feeding into a common downstream node.

\subsection{Construction Pipeline}

After defining the task representation, we explain how tasks evolve across rounds.
As shown in Figure~\ref{fig:main}, the pipeline starts from a simple solvable query $q_1$ and iterates Explorer, Formalizer, and Challenger in sequence until the stopping criterion is met.
In each round, the Explorer performs a rollout of the current query in the environment and produces a trajectory that includes the explored information.
The Formalizer parses the trajectory and updates the task graph $G$ and the corresponding rubrics.
The Challenger looks beyond the current task graph, identifies related directions that can harden the query, and turns them into the next harder query.
We describe the design of each component below.

\textbf{Explorer.}
In round $t$, the Explorer rolls out the query with tools over external sources and returns a trajectory
\[
\mathrm{Explorer}(q_t) \to traj_t,
\]
which provides traceable and scenario-relevant information for task evolution.
A single exploration rarely gathers enough information for a deep research task, because specialized knowledge is often long-tail, sparse, and cross-source.
Such knowledge becomes discoverable only when search intents already name the relevant terminology, concepts, or constraints, which our construction pipeline is designed to achieve.
For $t \geq 2$, the Challenger builds a harder $q_t$ from $G_{t-1}$ and $traj_{t-1}$ that incorporates knowledge accumulated across rounds.
This enables more precise search and drives the Explorer to uncover hard-to-find professional information hidden in the vast search space.

\textbf{Formalizer.}
The Formalizer turns the trajectory obtained by the Explorer into the DAG form defined above, and derives an aligned verifiable rubric set from it.
In each round, it reads $(q_t, G_{t-1}, traj_t)$ and performs
\[
\mathrm{Formalizer}(q_t, G_{t-1}, traj_t) \to (G_t, R_t),
\]
where there is no prior graph in the first round.
In the first round ($t=1$), the Formalizer extracts established evidence and analysis from $traj_1$ and organizes them into the first DAG $G_1$.
In later rounds ($t \geq 2$), it revises $G_{t-1}$ according to $q_t$ and $traj_t$ to produce $G_t$.
The principle of using $G_t$ to represent the solving process of $q_t$ is to keep the DAG sufficient for answering $q_t$ while keeping it minimal.
Sufficient means the $G_t$ covers what is needed for the open-ended answer required by $q_t$, with as many research aspects and as much supporting content as possible.
Minimal means that all information in $G_t$ should materially help answer $q_t$ and be mutually non-substitutable.
Therefore, the Formalizer first deletes nodes unrelated to $q_t$ and merges semantically equivalent nodes.
It then adds as much supported information from $traj_t$ as possible that falls within the scope of $q_t$ but is not yet in the graph, and finally constructs dependency edges among the remaining nodes.

Once the task graph is obtained, the Formalizer derives the rubric set $R_t = \{r_v\}_{v \in V_t}$ from $G_t$, where $V_t$ is the node set of $G_t$, and assigns one rubric $r_v$ to each node $v$. Each $r_v$ checks the execution quality of the step described by $desc_v$ via checkpoints $ckpt_v$. 
This design has two benefits.
First, $G_t$ already controls the sufficient and minimal coverage for solving $q_t$, so the derived rubric set $R_t$ evaluates exactly what $q_t$ requires and nothing beyond that.
Second, each checkpoint in $ckpt_v$ comes from the interaction between the agent and the environment and can be verified independently, making evaluation both traceable and judgeable. 
The Formalizer then assigns weights from leaf nodes upward on the DAG to reflect the different importance of research steps.
Let $L_t \subseteq V_t$ be the leaf set of $G_t$ and $Pa(v)$ be the parent set of node $v$.
Each leaf starts with the same initial weight
\[
\tilde{w}_v = \begin{cases} 1/|L_t|, & v \in L_t \\ 0, & \text{otherwise} \end{cases}
\]
Then, in reverse topological order, each node passes its weight equally to its parents,
\[
\tilde{w}_u = \tilde{w}_u + \frac{\tilde{w}_v}{|Pa(v)|}, \quad \forall u \in Pa(v),
\]
when $Pa(v) \neq \emptyset$ and $\tilde{w}_v > 0$.
Finally, the weights are normalized to sum to one, and rubric $r_v$ receives
\[
w_v = \frac{\tilde{w}_v}{\sum_{u \in V_t} \tilde{w}_u}.
\]
In this way, weights follow the DAG structure.
Prior steps that support more downstream conclusions accumulate more weight, while independent parallel aspects receive comparable weight through separate leaves.
At evaluation time, binary judgments are assigned to the checkpoints under each rubric $r_v$, with their satisfied fraction defining $s_v\in[0,1]$, and the score for task $q_t$ is then computed as $\sum_{v\in V_t} w_v s_v$.

Notably, we use the DAG structure to assign rubric weights, but not to check whether the model reproduces that structure.
This is because the same deep research task admits different solution paths due to its openness.
Therefore, in evaluation, the role of the rubric is to define what must be covered, not to require the tested model to reproduce the same solving path in the DAG.
Accordingly, we also control wording in the rubrics to evaluate final responses rather than fixed solving procedures and do not rigidly constrain retrieval sources, tool choice, or phrasing.

\textbf{Challenger.}
The Challenger evolves $q_t$ into a harder query $q_{t+1}$ while keeping it logically coherent and solvable.
In round $t$, it reads $(G_t, traj_t)$ and performs
\[
\mathrm{Challenger}(G_t, traj_t) \to q_{t+1}.
\]

Specifically, let $\mathcal{I}(\cdot)$ denote the information contained in a trajectory, graph, or query, and let $C_t$ be the selected expansion directions incorporated into $q_{t+1}$.
These directions come from explored but unused knowledge,
\[
C_t \subseteq \mathcal{I}(traj_t) \setminus \mathcal{I}(G_t).
\]
The next query then asks about both the used graph knowledge and these directions,
\[
\mathcal{I}(q_{t+1}) = \mathcal{I}(G_t) \cup C_t,
\]
making $q_{t+1}$ harder than $q_t$, which only required the coverage of $G_t$.
Because this expansion draws only on information already reached in exploration, $q_{t+1}$ stays relevant to the current query and remains solvable.

In our design, the Challenger can make the query harder along both width and depth.
Along width, it adds relevant aspects that lie more hops away from the current solved task.
Along depth, it reframes the current task into a harder one that requires longer retrieval or reasoning chains.
Besides, the next query hides detailed graph structure and evidence paths to encourage the Explorer to explore new space and provide sufficient information for deep research.
To keep the Challenger from easing the task when wording $q_{t+1}$, we mask node checkpoints $ckpt_v$ to prevent conclusions from being exposed on the query.

\textbf{Stopping Criterion.}
We stop task evolution when both the nodes and edges of the DAG remain unchanged.
Let $G_t=(V_t,E_t)$ denote the DAG after round $t$.
The stopping time is
\begin{equation}
T=\min\left\{t\geq 3:
G_t=G_{t-1}=G_{t-2}
\right\}.
\end{equation}
It requires the complete graph to stay unchanged for two consecutive rounds, because a single unchanged round may be accidental: the Explorer may bring back no new information in that rollout, the Formalizer may add nothing to the DAG, or the Challenger may find no expansion directions.

\subsection{Query Design for Capability Probing}

From the final task, we construct three query types that share the same rubric set $R_T$ and task DAG $G_T$ and differ only in the user input shown to the model.
They test complementary deep research abilities by varying how explicitly the solving clues are disclosed in the user input.
\textbf{The query with hints $q_T^{\mathrm{h}}$} is the final Challenger query $q_T$ itself, which keeps analysis dimensions, and solving constraints.
It tests information processing under rich clues, including retrieval organization and cross-source integration of noisy evidence.
\textbf{The query without hints $q_T^{\mathrm{o}}$} is a shorter everyday question derived from $q_T^{\mathrm{h}}$ that drops analysis dimensions and solving constraints. 
It tests decomposition and planning under hidden constraints, including identifying research coverage, decomposing subquestions, and planning execution order.
\textbf{The assigned-topic query $q_T^{\mathrm{a}}$} is a declarative research title generated from the DAG $G_T$, stating the research subject and downstream conclusions rather than asking a question.
It tests argument formation under a directional topic only, including inferring the required information work from one sentence and building a coherent and persuasive answer.
\section{Benchmark Evaluation}

\begin{table}[t]
    \centering
    \footnotesize
    \begin{tabular}{lrrr}
    \toprule
    Metric & Min & Max & Mean \\
    \midrule
    Total tool calls & 32 & 1854 & 264.96 \\
    Evolution rounds & 4 & 58 & 14.19 \\
    DAG depth & 2 & 28 & 6.21 \\
    DAG breadth & 2 & 29 & 10.98 \\
    DAG nodes & 12 & 40 & 20.85 \\
    Checkpoints & 52 & 183 & 104.85 \\
    \bottomrule
    \end{tabular}
    \caption{Statistics of the constructed benchmark.}
    \label{tab:retained}
\end{table}

\subsection{Experimental Setup}

We select queries from NQ-Open \citep{nqopen} as seed queries, an open domain QA corpus derived from Wikipedia whose cases are simple factual questions with reference answers. 
We construct the benchmark with GPT-5.5 serving as the Explorer, Formalizer, and Challenger. 
After task evolution stops, we keep successfully converged samples (96.3\% success rate, with failure reasons summarized in the Appendix) and retain the 500 tasks with the most DAG nodes to ensure sufficient difficulty.

We test ten popular models for evaluation:
DeepSeek-V4-Pro~\citep{deepseekv4}, Qwen3.7-Max~\citep{qwen37max}, GLM-5.2~\citep{glm52}, GPT-5.6~Terra~\citep{gpt56terra}, Claude~Sonnet~5~\citep{claudesonnet5}, Gemini~3.5~Flash~\citep{gemini35flash}, MiniMax-M2.5~\citep{minimaxm25}, Qwen3.6-Flash~\citep{qwen36flash}, Qwen3-32B~\citep{qwen3technicalreport}, and Qwen3-14B~\citep{qwen3technicalreport}.
The evaluator is Qwen3.7-Max.
We test under two settings.
\textbf{Agent mode} uses the same tool setting as benchmark construction.
\textbf{Model mode} relies only on internal model knowledge for reasoning.
Additional implementation details and experimental settings are provided in the Appendix to facilitate reproducibility.

\subsection{Benchmark Overview}


\begin{figure}[t]
\centering
\includegraphics[width=0.8\linewidth]{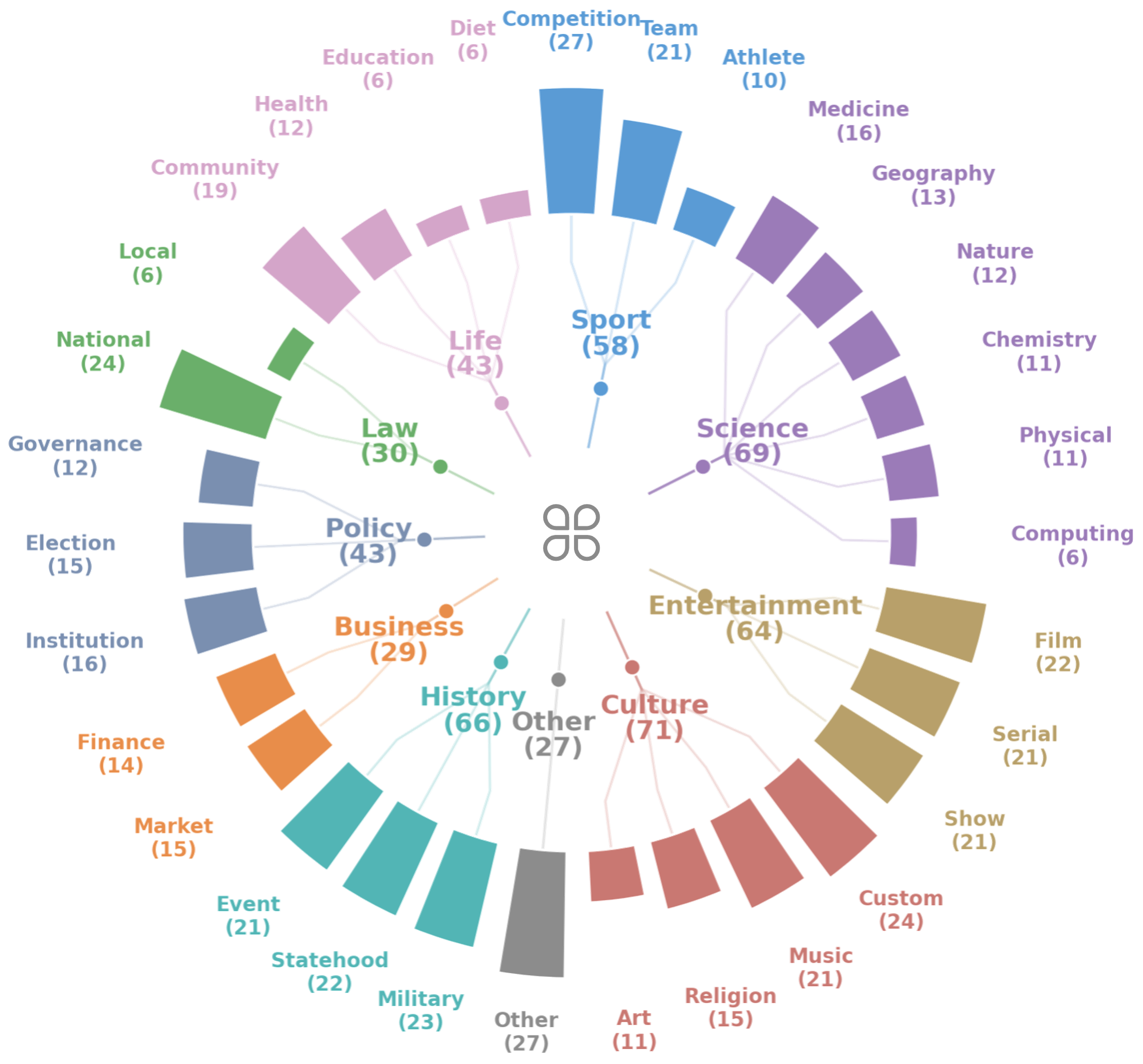}
\caption{Topic distribution of tasks in the benchmark.}
\label{fig:topics}
\end{figure}

\begin{table}[t]
\centering
\footnotesize
\setlength{\tabcolsep}{0pt}
\begin{tabular}{p{0.56\linewidth}@{\hspace{2.5pt}}r@{\hspace{2.5pt}}r@{\hspace{2.5pt}}r}
\toprule
Question & Poor & Medium & Good \\
\midrule
Is $q_T^{\mathrm{h}}$ meaningful? & 0 & 0 & 100 \\
Is $q_T^{\mathrm{h}}$ hard enough? & 0 & 2 & 98 \\
Does $q_T^{\mathrm{o}}$ keep the original meaning? & 0 & 2 & 98 \\
Is $q_T^{\mathrm{o}}$ concise enough? & 0 & 1 & 99 \\
Does $q_T^{\mathrm{a}}$ match the main topic? & 0 & 0 & 100 \\
Is $G_T$ correct? & 0 & 0 & 100 \\
Is $R_T$ discriminative? & 0 & 0 & 100 \\
Does $R_T$ cover the important points? & 0 & 0 & 100 \\
Is the strictness of $R_T$ appropriate? & 0 & 0 & 100 \\
Are the weights in $R_T$ reasonable? & 0 & 1 & 99 \\
Are the checkpoints in $R_T$ verifiable? & 0 & 0 & 100 \\
\bottomrule
\end{tabular}
\caption{Quality review by human on 100 tasks.}
\label{tab:review}
\end{table}

\begin{table*}[t]
\centering
\footnotesize
\begin{tabular}{lcccccc}
\toprule
Model & \multicolumn{3}{c}{Agent mode} & \multicolumn{3}{c}{Model mode} \\
\cmidrule(lr){2-4}\cmidrule(lr){5-7}
& $q_T^{\mathrm{h}}~(\mathcal{E}{+}\mathcal{A})$ & $q_T^{\mathrm{o}}~(\mathcal{E}{+}\mathcal{A})$ & $q_T^{\mathrm{a}}~(\mathcal{E}{+}\mathcal{A})$ & $q_T^{\mathrm{h}}~(\mathcal{E}{+}\mathcal{A})$ & $q_T^{\mathrm{o}}~(\mathcal{E}{+}\mathcal{A})$ & $q_T^{\mathrm{a}}~(\mathcal{E}{+}\mathcal{A})$ \\
\midrule
GPT-5.6~Terra & \textbf{0.86}~(\underline{0.41}{+}\underline{0.45}) & 0.67~(0.30{+}\underline{0.37}) & 0.70~(0.32{+}\underline{0.38}) & \textbf{0.68}~(0.30{+}\underline{0.38}) & \textbf{0.54}~(\underline{0.23}{+}\underline{0.31}) & 0.56~(0.25{+}\underline{0.31}) \\
GLM-5.2 & 0.80~(0.39{+}0.41) & \textbf{0.68}~(\underline{0.31}{+}\underline{0.37}) & 0.69~(0.34{+}0.35) & 0.65~(0.30{+}0.35) & 0.52~(\underline{0.23}{+}0.29) & \textbf{0.57}~(\underline{0.26}{+}\underline{0.31}) \\
DeepSeek-V4-Pro & 0.83~(\underline{0.41}{+}0.42) & \textbf{0.68}~(\underline{0.31}{+}\underline{0.37}) & \textbf{0.73}~(\underline{0.35}{+}\underline{0.38}) & 0.62~(0.28{+}0.34) & 0.48~(0.22{+}0.26) & 0.53~(0.24{+}0.29) \\
Claude~Sonnet~5 & 0.80~(0.38{+}0.42) & 0.67~(0.30{+}\underline{0.37}) & 0.71~(0.33{+}\underline{0.38}) & 0.62~(0.28{+}0.34) & 0.47~(0.20{+}0.27) & 0.51~(0.23{+}0.28) \\
Qwen3.7-Max & 0.83~(0.40{+}0.43) & 0.67~(\underline{0.31}{+}0.36) & 0.71~(0.34{+}0.37) & 0.60~(0.27{+}0.33) & 0.47~(0.21{+}0.26) & 0.48~(0.21{+}0.27) \\
Gemini~3.5~Flash & 0.74~(0.34{+}0.40) & 0.57~(0.25{+}0.32) & 0.63~(0.29{+}0.34) & 0.67~(\underline{0.31}{+}0.36) & 0.52~(\underline{0.23}{+}0.29) & 0.54~(0.25{+}0.29) \\
MiniMax-M2.5 & 0.67~(0.30{+}0.37) & 0.42~(0.18{+}0.24) & 0.55~(0.26{+}0.29) & 0.46~(0.20{+}0.26) & 0.31~(0.14{+}0.17) & 0.35~(0.17{+}0.18) \\
Qwen3.6-Flash & 0.53~(0.23{+}0.30) & 0.40~(0.17{+}0.23) & 0.43~(0.20{+}0.23) & 0.44~(0.19{+}0.25) & 0.28~(0.12{+}0.16) & 0.38~(0.17{+}0.21) \\
Qwen3-32B & 0.36~(0.16{+}0.20) & 0.23~(0.11{+}0.12) & 0.23~(0.11{+}0.12) & 0.30~(0.13{+}0.17) & 0.18~(0.08{+}0.10) & 0.18~(0.09{+}0.09) \\
Qwen3-14B & 0.35~(0.15{+}0.20) & 0.21~(0.09{+}0.12) & 0.21~(0.10{+}0.11) & 0.27~(0.11{+}0.16) & 0.16~(0.07{+}0.09) & 0.16~(0.08{+}0.08) \\
\bottomrule
\end{tabular}
\caption{Performance on the benchmark. \textbf{Bold} denotes the best overall score and \underline{underline} the best $\mathcal{E}$/$\mathcal{A}$ score in each column.}
\label{tab:models}
\end{table*}

\begin{figure*}[!t]
\centering
\includegraphics[width=0.99\textwidth]{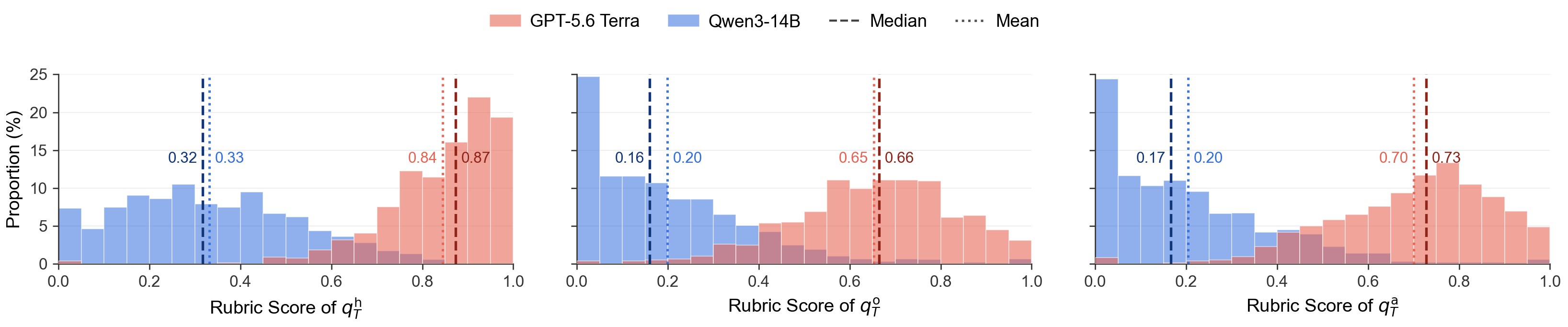}
\caption{Score distributions of the strongest and weakest models on $q_T^{\mathrm{h}}$ (left), $q_T^{\mathrm{o}}$ (middle), and $q_T^{\mathrm{a}}$ (right).}
\label{fig:fuzz_dist}
\end{figure*}

Table~\ref{tab:retained} summarizes tool use, evolution rounds, and DAG scale for the benchmark.
On average, evolving a simple query into a deep research task takes 14.19 rounds and 264.96 tool calls, and each constructed task has 104.85 checkpoints, reflecting the difficulty of our benchmark. We further summarize topic distribution for the benchmark, and Figure~\ref{fig:topics} shows counts by coarse and fine topic.
The 500 tasks cover common open domain retrieval themes rather than one narrow domain.
These statistics show that the benchmark contains domain-relevant, professionally deep research tasks with verifiable rubrics evolved from simple queries across diverse topics.

To assess the quality of the constructed data, three senior researchers familiar with retrieval and deep research tasks spent a total of 182 hours reviewing 100 tasks randomly sampled from the benchmark.
They rated each of eleven questions for every task as good, medium, or poor, following the criteria in the Appendix. 
Following majority vote (Fleiss' $\kappa=0.81$), Table~\ref{tab:review} shows no poor ratings and only a few medium ones, indicating the high usability of the constructed data.
Notably, all $10{,}695$ checkpoints that encode task knowledge were manually checked as reliable.

\subsection{Baseline Performance}

Table~\ref{tab:models} reports overall scores of ten models across the three query types under agent and model modes, each decomposed as $\mathcal{E}{+}\mathcal{A}$ with average full marks of 0.52 and 0.48 for evidential and analytical checkpoints, and Figure~\ref{fig:fuzz_dist} shows the score distributions of the strongest and weakest models on each query type.
Across all three query types and both modes, the benchmark shows strong discrimination and clearly separates stronger models from weaker ones.
Both across models and within each model, scores spread continuously rather than clustering at a few discrete levels, showing that the rubrics provide fine-grained graded satisfaction rather than binary judgments.
Compared with $q_T^{\mathrm{h}}$, agent-mode scores drop by about 0.16 on $q_T^{\mathrm{o}}$ and 0.12 on $q_T^{\mathrm{a}}$, showing agents’ reliance on explicit research guidance.

\begin{table}[t]
\centering
\setlength{\tabcolsep}{1.5pt}
\renewcommand{\arraystretch}{1.08}
\footnotesize
\begin{tabular}{@{}lcccccc@{}}
\toprule
& \multicolumn{2}{c}{Manual}
& \multicolumn{3}{c}{Semi-automatic}
& Automatic \\
\cmidrule(lr){2-3}\cmidrule(lr){4-6}\cmidrule(lr){7-7}
Metric
& RR
& DEER
& DR-I
& DR-II
& Arena
& Ours \\
\midrule
Pearson $r$
& \na
& $0.73$
& $0.6024$
& \na
& $0.79$\,/\,$0.76$
& $0.899$ \\
Spearman $\rho$
& \na
& $0.71$
& $0.5912$
& \na
& $0.84$\,/\,$0.81$
& $0.902$ \\
Macro-F1
& $0.73$
& \na
& \na
& \na
& \na
& $0.856$ \\
Accuracy
& \na
& \na
& \na
& $0.9175$
& \na
& $0.862$ \\
F1
& \na
& \na
& \na
& $0.8957$
& \na
& $0.834$ \\
Agreement
& \na
& $0.84$
& $0.7133$
& \na
& \na
& $0.871$ \\
\bottomrule
\end{tabular}
\caption{Human--machine consistency across benchmarks.}
\label{tab:task-level-human-agreement}
\end{table}

\begin{table*}[t]
\centering
\captionsetup{skip=6pt}
\begin{minipage}[t]{0.32\textwidth}
\centering
\footnotesize
\setlength{\tabcolsep}{3pt}
\begin{tabular}{@{}lccc@{}}
\toprule
Judge\vphantom{$R_{10}$} & $q_T^{\mathrm{h}}$ & $q_T^{\mathrm{o}}$ & $q_T^{\mathrm{a}}$ \\
\midrule
Qwen3.7-Max (J1) & 0.916 & 0.884 & 0.893 \\
Qwen3.7-Max (J2) & 0.916 & 0.888 & 0.893 \\
Qwen3.7-Max (J3) & 0.916 & 0.893 & 0.894 \\
GLM-5.2 & 0.901 & 0.884 & 0.893 \\
DeepSeek-V4-Pro  & 0.897 & 0.895 & 0.902 \\
\bottomrule
\end{tabular}
\caption{Judge correlation with human.}
\label{tab:rubric_consist}
\end{minipage}
\hfill
\begin{minipage}[t]{0.67\textwidth}
\centering
\footnotesize
\setlength{\tabcolsep}{1pt}
\begin{tabular}{@{}l*{15}{c}@{}}
\toprule
Metric & $R_1$ & $R_2$ & $R_3$ & $R_4$ & $R_5$ & $R_6$ & $R_7$ & $R_8$ & $R_9$ & $R_{10}$ & $R_{11}$ & $R_{12}$ & $R_{13}$ & $R_{14}$ & $R_{15}$ \\
\midrule
Tasks & 500 & 500 & 500 & 500 & 499 & 492 & 476 & 447 & 398 & 338 & 302 & 253 & 225 & 204 & 172 \\
Tool calls & 3.9 & 12.6 & 16.0 & 17.7 & 18.7 & 19.5 & 20.0 & 20.1 & 20.6 & 20.6 & 20.9 & 22.0 & 21.7 & 22.5 & 23.3 \\
Depth & 2.5 & 3.8 & 4.4 & 4.9 & 5.2 & 5.5 & 5.7 & 5.9 & 5.9 & 6.2 & 6.3 & 6.6 & 6.7 & 6.9 & 7.0 \\
Nodes & 4.6 & 8.6 & 11.6 & 13.6 & 15.3 & 16.7 & 18.0 & 18.9 & 20.0 & 21.1 & 22.0 & 22.9 & 23.8 & 24.8 & 25.2 \\
Checkpoints & 13.6 & 33.5 & 50.7 & 63.0 & 73.6 & 81.3 & 88.3 & 93.4 & 98.8 & 104.4 & 109.7 & 113.1 & 116.7 & 120.9 & 123.2 \\
\bottomrule
\end{tabular}
\caption{Per-round metrics during benchmark construction.}
\label{tab:per_iter}
\end{minipage}
\end{table*}

\begin{figure}[!t]
\centering
\includegraphics[width=0.48\textwidth]{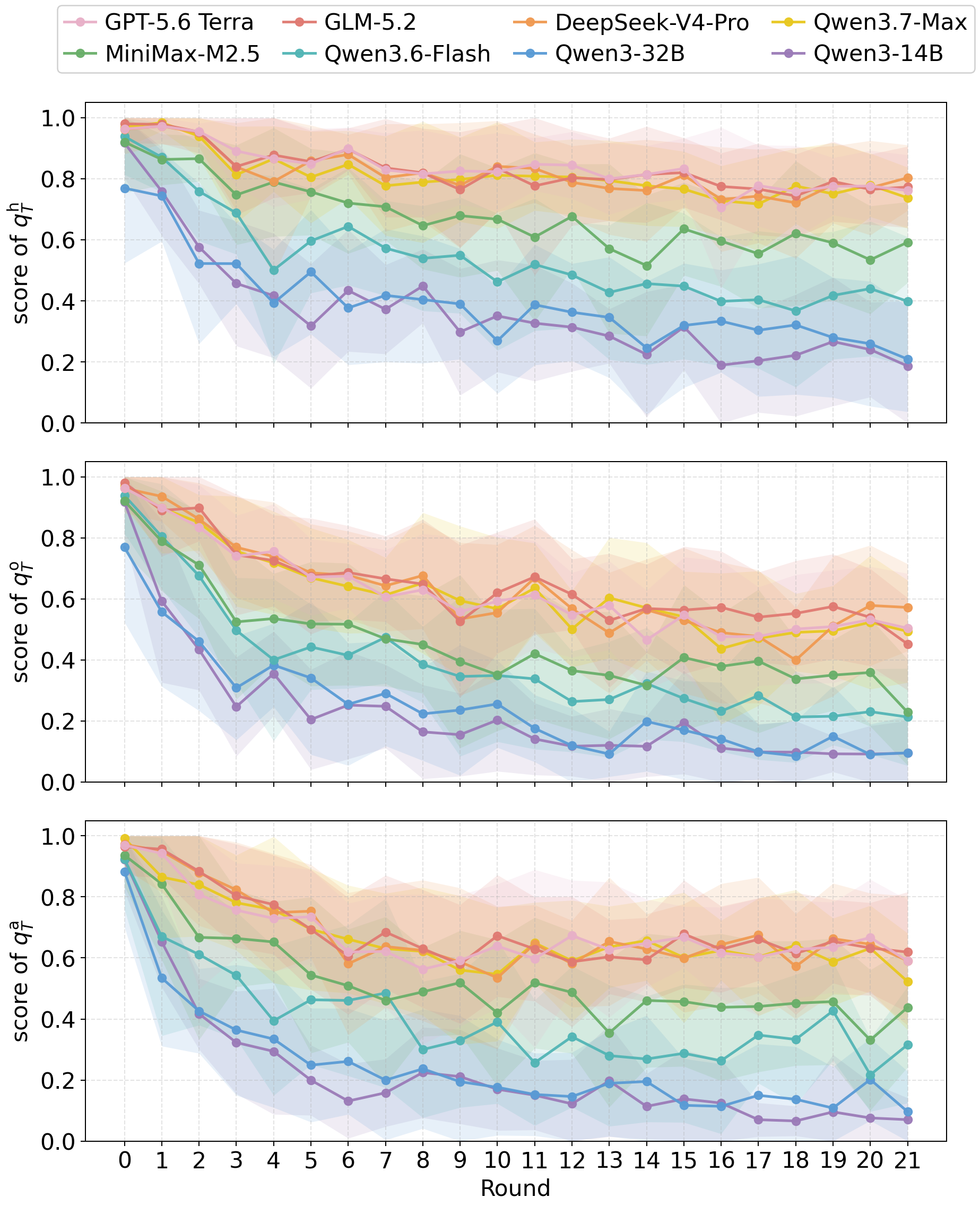}
\caption{Performance across evolution rounds.}
\label{fig:evolution_agg}
\end{figure}


\begin{table*}[!t]
\centering
\footnotesize
\setlength{\tabcolsep}{0pt}
\begin{tabular}{@{}p{0.38\textwidth}@{}p{0.55\textwidth}@{}>{\centering\arraybackslash}p{0.08\textwidth}@{}}
\toprule
Simple query & Main evolution directions & Similarity \\
\midrule
\emph{who played fred claus in the 2007 movie} & cast credits, movie reviews, character story, family roles & $0.27$ \\
\emph{what was the name of the first mission to the moon} & first-mission definition, Apollo focus & $0.21$ \\
\emph{where do they store dead bodies in the hospital} & legal rules, morgue, overcrowding, hospital transfer, temporary storage & $0.18$ \\
\emph{who has the most followers on the twitter} & leadership reasons, recommendation algorithms, cross-platform comparison & $0.17$ \\
\emph{who owns the rights to rubik's cube} & ownership history, puzzle competition, cross-country rights & $0.22$ \\
\bottomrule
\end{tabular}
\caption{Diversity statistics over repeated construction from five simple queries.}
\label{tab:repeat}
\end{table*}

Across settings, the benchmark tests different capabilities and exposes distinct model strengths and weaknesses. 
Here are some interesting observations.
GPT-5.6~Terra, GLM-5.2, DeepSeek-V4-Pro, Claude~Sonnet~5, and Qwen3.7-Max show strong performance in the agent mode, while Gemini~3.5~Flash is weaker with tools but can surpass most of them using internal knowledge alone.
While evidential scores stay close among these frontier models, GPT-5.6~Terra leads clearly on analytical checkpoints, and this stronger reasoning helps it top the overall ranking.
Relative to $q_T^{\mathrm{h}}$, weaker hints shift the strong--weak gap from being roughly balanced across evidential and analytical checkpoints toward being driven more by analytical reasoning, suggesting that smaller models struggle more with research abilities such as decomposing a broad goal into subtasks and combining evidence into a coherent answer.
MiniMax-M2.5 shows a clear cliff on $q_T^{\mathrm{o}}$, whereas $q_T^{\mathrm{a}}$ does not produce a comparable drop, suggesting that its weakness lies in uncovering the hidden goal of a query without hints, rather than in conducting research once the topic is clear.

Notably, removing tools causes large drops for all ten models on all query types: overall scores fall by 0.14 on average, with mean drops of 0.07 on both evidential and analytical checkpoints.
This shows that our tasks encode long-tail information that models rarely store in pretraining and must instead recover through deep search.
This suggests the potential to extend our construction pipeline to professional domains, where it could uncover in-depth knowledge and evolve simple queries into verifiable deep research tasks.

We also evaluate the stability of the rubrics under rejudging on the 100 manually reviewed tasks.
As shown in Table~\ref{tab:rubric_consist}, whether the same model is used for repeated judging or different models are used as judges, the scores all show high Pearson correlation with human ratings.
Table~\ref{tab:task-level-human-agreement} further situates these results against other deep research benchmarks that use task-level rubrics and report human--machine consistency.
Our automatically constructed rubrics remain competitive with, and often stronger than, expert-authored or semi-automatic alternatives on the comparable correlation and agreement measures.
This indicates that task-specific rubrics can support reliable and human-aligned evaluation of open-ended deep research tasks.

    

\subsection{Construction Analysis}

To characterize how tasks evolve during construction, Table~\ref{tab:per_iter} aggregates construction metrics over the first 15 rounds, which cover the mean evolution round $14.19$.
Overall, task evolution follows a deepen-then-broaden pattern as long-tail 
domain knowledge is progressively uncovered. 
DAG depth rises quickly from 2.5 in round 1 to 5.2 in round 5, then more slowly toward about 7.0, showing that construction first extends reasoning chains.
Afterwards, node count continues to grow from 15.3 to 25.2, so later gains mainly come from broader parallel research aspects rather than longer chains.
Explorer tool calls and checkpoints both grow smoothly, indicating that tasks steadily become harder as construction proceeds.

We further evaluate 20 randomly selected cases at every round to check whether difficulty rises from the solving side.
Figure~\ref{fig:evolution_agg} reports mean scores over the evolution rounds reached by all cases. Each line is the mean score of one model over ten cases and the shaded band is the standard deviation across cases.
Scores for all three query types decline overall as $t$ increases, confirming that task difficulty rises steadily over evolution. 
The $q_T^{\mathrm{h}}$ setting drops more gently and often stays in the mid to high range because it reflects solvability under rich information.
$q_T^{\mathrm{o}}$ and $q_T^{\mathrm{a}}$ drop more sharply.
Their round means fall from about 0.93 and 0.95 to about 0.29 and 0.40 with larger variance.
Relative model ranking stays stable across most rounds, which shows that difficulty increases without breaking the capability gradient. 
We examine whether scores on converged tasks are associated with structural metrics.
None of DAG depth, node count, round count, and tool-call count shows a stable association with scores.
These results provide empirical support that our stopping criterion identifies task-specific saturation points across different structural scales, with final task difficulty not systematically determined by DAG size.

\subsection{More Analysis}

To test the diversity of constructed tasks, we randomly select five simple queries and evolve each one ten times independently.
Table~\ref{tab:repeat} reports the results, and we use Jaccard similarity among the final queries after construction to measure their lexical overlap.
The mean Jaccard similarity ranges from 0.17 to 0.27, which shows that queries diverge along multiple research directions rather than through minor wording changes.
Although every seed yields at least one deep research task that humans judge as sufficiently hard, we observe that evolution quality depends on the scope of the seed query.
Specifically, fact-seeking seeds usually stop earlier and yield multi-source search tasks, while more open seeds run longer and yield tasks that emphasize integrative analysis. 
More case analyses highlighting the strengths and weaknesses of our benchmark are provided in the Appendix.

\section{Conclusion}

We present a verifiable benchmark of 500 deep research tasks with source-grounded rubrics, constructed from a Wikipedia QA corpus without expert-authored tasks or pre-written human materials.
To build the benchmark, an iterative Explorer--Formalizer--Challenger pipeline expands simple queries into DAGs of atomic steps with verifiable checkpoints, so that the query, graph, and rubrics evolve together.
Experiments show a clear capability gradient across models and query types and demonstrate that fact-grounded pointwise rubrics provide fine-grained, human-aligned discrimination.
One limitation of our work is the relatively high construction cost resulting from the use of a frontier model during task evolution, which we view as a necessary investment in building a trustworthy benchmark for the community.

\newpage
\bibliography{aaai2026}

\clearpage
\onecolumn
\begin{center}
{\LARGE\bf Appendix}
\end{center}
\vspace{0.5em}
\section{Implementation Details and Experimental Settings}
\label{app:implementation}

\subsection{Prompt Templates}

The following are the full construction and query-transformation prompts.

\subsubsection{Explorer Prompt}
\mbox{}\\[-0.6em]
\begin{lstlisting}
# Your role
- You are a general problem-solving assistant: decompose the user need into executable information-retrieval and multi-perspective analysis tasks, call available tools to obtain evidence, then answer with a complete deep research report.
- Grounding in evidence is the first principle: your answers and reasoning must be supported by evidence; do not answer without retrieval.

# Current date
- {current_time}

# Output format requirements
- All answer content must be fully enclosed in <answer></answer> tags: <answer>...your complete answer...</answer>
- Answer in English; do not insert images.

# Core workflow (must follow)
- Answer based on tool_response; do not give conclusions when evidence is insufficient.
- If tools return empty / failed / conflicting results, adjust the retrieval strategy (change keywords, sources, time range, broaden/narrow scope) and call tools again.

# Tool calling strategy (important)
- **Prefer parallel calls**: when multiple mutually independent information sources or dimensions are needed, query them in parallel in the same round (e.g., different sources, years, or institutional standards).
- **Sequential call scenarios**: use sequential calls only when a later call depends on the previous return (e.g., obtain entity IDs/list first, then query details).
- **Deduplication and restraint**: avoid repeatedly calling the same tool with the same parameters; before each call, clarify the specific point to verify (do this mentally; do not expose instructions in the answer).

# Tool usage suggestions
- **Start with `google_search`**: turn the user question into 2-5 keyword/qualifier sets (entity names, time range, standard terms, synonyms), and prefer parallel retrieval.
- **Then use `web_scrape`**: for key webpage URLs returned by `google_search` that you need to cite, scrape the full text one by one and use scraped results as evidence.
- **Failure fallback**: if `web_scrape` fails or content is incomplete, scrape other source links on the same topic, or adjust `google_search` keywords and time filters and retry.

# Evidence and traceability requirements
- For each key conclusion, provide corresponding evidence (from which data item / which original excerpt).
- Clearly mark the time range / effective date of information; for "latest / current / recent", you must state "as of {current_time} or the date in the tool response".

# Current question (query)
- {query}
\end{lstlisting}

\subsubsection{Formalizer Prompt}
\mbox{}\\[-0.6em]
\begin{lstlisting}
You are an assistant that revises the input mermaid_old using the given query and trajectory so that mermaid_new shows the complete deep research process needed to answer the query, and then derives a rubric from mermaid_new.

## Known inputs
- query:
{query}

- mermaid_old:
{mermaid_old}

- trajectory:
{trajectory}

# Your workflow
First, depending on whether mermaid_old is empty, follow Plan 1 or Plan 2 to produce mermaid_new. Mermaid content requirements are specified below.

[Plan 1: mermaid_old is empty]
Extract key nodes, logical relations, and node information from the trajectory, and directly generate a compliant `mermaid_new`.

[Plan 2: mermaid_old is not empty]
Revise mermaid_old to fit the current query and obtain `mermaid_new`:
- Delete unrelated information: based on the given query, delete information that appears in mermaid_old but is irrelevant to the query's final goal. Delete only nodes that are truly out of scope; do not delete evidential or analytical nodes that still serve the query merely to make the graph look shorter.
- Add related information: the trajectory may contain information that serves the current query but is not yet covered by mermaid_new; add that information, staying as concrete and complete as possible without going beyond the query scope.
- Merge equivalent nodes:
  - Merge **only when** several nodes point to the **same fact, same object, and same verifiable conclusion** under different wording or retrieval channels (e.g., multiple sources only confirm that Vince Vaughn plays Fred Claus).
  - **Do not** merge multiple independent key events that can be retrieved and scored separately, even if they share a theme. For example, different victims, scenes, timeline anchors, or narrative functions of multiple killings/conflicts should remain separate **evidential** nodes. If synthesis is needed, add a downstream **analytical** node.
  - **Analytical** nodes may be reorganized or rephrased, but must not be merged in a way that hides independent evidential nodes that should be kept.
- Adjust the mermaid so it is more reasonable and clear:
  - Adjust structure: ensure logical relations among nodes are correct and that the whole graph reflects the intrinsic information logic of deep research for solving the query (do not reorder by the trajectory's execution order).
  - Adjust node information: keep node fields consistent. For unchanged nodes from mermaid_old, keep node_type and check_points unchanged when possible; source_type and source_note may be updated when needed, but prefer leaving them unchanged when unnecessary.

**Important:** All revisions aim to make mermaid_new better suited to completing the query than mermaid_old. Deletion removes out-of-scope nodes; addition fills missing information; merging and restructuring improve clarity. If mermaid_old already satisfies the requirements above, mermaid_new may be identical to mermaid_old.

Then, from mermaid_new, generate a rubric used to judge how other models perform on the same query:
- Produce one criterion for every node in mermaid_new. Each criterion includes:
  - `node_id`: must match the corresponding `node_id` in `mermaid_new.nodes` **exactly** (case-sensitive). If the structure node is `A`, write `"A"`; do not use aliases such as `node_a`.
  - `description`: reuse the action+object text of that node in structure to state what is being judged.
  - `check_points`: reuse the node's check_points, splitting the description into concrete, independently verifiable checkpoints that are mechanically judgeable; avoid subjective words such as comprehensive / deep / high-quality. Include substantive evidential content, e.g., not merely "includes film criticism", but "includes criticism about the O. Henry-style ending".
- Criterion content must come from the nodes and may be lightly adapted to how each node serves the query, but must not invent facts. For substitutable evaluation content, do not be overly strict:
  - Unless the query explicitly requires it, do not require specific tools, search engines, URLs, webpage sources, tool calls, or retrieval steps, because any search path may obtain the same information; only check whether the result is correct. E.g., rewrite "Retrieved from Wikipedia that Andy Lau stars in Infernal Affairs" as "Andy Lau stars in Infernal Affairs".
  - Unless a term must appear, do not require specific keywords or fixed phrasing; judge meaning, not wording.
  - Unless a specific example must appear, allow different but reasonable examples. If an example is needed, write "provide at least one relevant instance or evidence".
  - Unless the query explicitly requires comparing source credibility, do not write rigid items such as "cross-verify across how many databases / prioritization workflow"; evaluate evidence sufficiency and logical consistency of the final conclusion.
  - Do not mention mermaid structural information (e.g., "depends on node A", or "node A's output is node B's input"); for the evaluated agent, mermaid_new does not exist.

# Mermaid requirements
The mermaid is the internal logic of the deep research task for answering the given query: node content reflects information, and edges reflect logical dependence. It is independent of the trajectory's tool-call order and concrete operations; the trajectory only supplies information, of which the mermaid uses a relevant subset. Fields and requirements:

- `mermaid.structure` is a DAG mermaid flowchart (no back-edges or cycles), containing only node content and connections:
  - Node content: each node describes one concrete **action + object** for solving the task, clear and specific, without writing the result, but with a node id and description, e.g., `A["Search Beijing weather"]`, `B["Estimate Zijin Mining valuation"]`.
    - Node granularity: whether evidential or analytical, one node corresponds to at most a single operation--e.g., one searched object, or one analytic question. Do not merge multiple operations. In particular, do not create global overview nodes such as `A["Analyze that the overall task needs surveying xxx then writing xxx"]` or `A["Identify the task definition: explain the follower gap on platform X in 2026 and how the lead was built over time"]`; split them into the concrete operational nodes.
    - Node count: unlimited; 1 to 100 nodes are all acceptable. Prefer completeness without duplication, fabrication, or omission.
  - Node connections: reflect actual information flow; multiple sources or sinks are allowed. Organize edges by logical dependence:
    - Derivation = vertical organization: a downstream node directly uses upstream information. Use vertical organization only when the downstream node is definitively derived from the upstream node, otherwise treat as parallel:
      - E.g., `B["Query Ping An Bank shareholder-count change"]` requires knowing which stock, so `A["Look up the stock name for 000001"] --> B["Query Ping An Bank shareholder-count change"]` is valid vertical organization.
      - E.g., `C["Summarize Vince Vaughn's pre-2007 comedy roles and performance style"]` presupposes `B["Confirm the actor who plays Fred Claus"]` (B's checkpoint: `["Retrieval confirms Vince Vaughn plays Fred Claus"]`), so `B --> C` is valid vertical organization.
    - Complementary (parallel) = horizontal organization: parallel nodes are different aspects of the same task and can be completed independently even if they share an object.
    - Connections depend on information dependence, not solving order. E.g., if the query asks both "How does casting in the new Lion King serve 'filling in prehistory' and 'continuing the story'?" and "How would answers differ if the new Lion King means the 2019 film?", the latter does not depend on the former, so do not organize them as A --> B.

- `mermaid.nodes` stores detailed information for every node in structure; **every** structure node needs one `mermaid.nodes` record:
  - `node_id`: must match the ID in `mermaid.structure` **exactly** (case-sensitive). If structure has `A["Search Beijing weather"]`, write `"A"`; do not lowercase, add `node_` prefixes, or use aliases.
  - `node_type`: must be exactly `"evidential"` or `"analytical"`.
    - evidential: information with a clear source or evidence that can stand without logical derivation; only information from `tool_response` may be labeled evidential.
    - analytical: information obtained by analyzing existing information (causal, inductive, comparative, predictive, etc.), still non-fabricated and originating from content already in the trajectory. **An analytical node must have at least one in-edge.**
  - `check_points`: the **concrete results/conclusions** that corroborate the node content; at least one per node, multiple allowed. Examples:
    - Evidential node `A["Query Zijin Mining enterprise value EV"]`: `["EV is about CNY 436 billion"]`
    - Analytical node `B["Compute the pressure on an object"]`: `["Substitute F=100N and S=2 m^2", "Apply p=F/S", "Obtain p=50Pa"]`
    - Analytical node `C["Analyze reasons for gold price rise"]`: `["Fed rate cuts weakened the USD and lifted gold", "Severe inflation reduced purchasing power", "Geopolitical conflict increased safe-haven demand for gold"]`
    - Analytical node `D["Analyze differences between EVs and fuel cars"]`: `["EVs use electricity while fuel cars rely on gasoline or diesel", "EVs usually have lower operating cost", "EVs are generally cleaner", "Fuel cars are often easier to refuel for long trips", "EVs tend to be more software-driven while fuel cars have a more mature conventional stack"]`
  - `source_type`: exactly one of:
    - `previous_mermaid`: from a node in mermaid_old (unchanged / no new node information this round).
    - `tool_call_response`: new content from this round's trajectory tool responses.
    - `query_or_reason_or_answer`: new content from this round's trajectory query / reason / final answer.
  - `source_note`: short source + evidence description (must be locatable in the input; no fabrication):
    - if `previous_mermaid`, write `from_mermaid_old` plus the inheritance basis.
    - if `tool_call_response`, write `tool_call_id=<id>` plus tool name/keywords and key evidence.
    - if `query_or_reason_or_answer`, specify fragment location (e.g., `query` / `reason` / `final_answer`) plus key evidence.

# Special notes
- **Out-of-scope nodes forbidden**: every node and checkpoint must directly or indirectly serve the query's final goal; delete out-of-scope information.

- Node content must be concrete and explicit: unless prior node content/checkpoints already name the referent, do not use pronouns such as "it", "this", or "that". E.g., rewrite `A["Identify this work as an allegory of freedom"]` as `A["Identify The Tale of Peter Rabbit as an allegory of freedom"]`.

- **Meaningless nodes forbidden**:
  - Any failed / empty result must not become a node, including request failure, scrape failure, unrendered placeholders, empty results, anti-bot blocks, timeouts, etc. Such failures may only be used internally to judge that a source is unavailable.
  - Reduce nodes such as "re-check / verify again / confirm via another channel". If answers differ, decide whether they concern the same node (merge) or distinct nodes (split). If a real conflict remains, verify twice which source is more authoritative; prefer omission over adding wrong information.
  - Merge repeated verification of the **same fact**. E.g., IMDb / Wikipedia / other lookups that all only confirm the same actor for Fred Claus should merge into one node. This is **not** merging distinct events/objects/scenes into one summary node; keep independent key-event nodes unless they truly verify the same conclusion.
  - **No horizontal compression of evidential nodes**: do not fold multiple distinct key events (different victims, executions/conflicts, historical crimes, climax scenes) into one "summary" evidential node; such synthesis belongs in an **analytical** node.
  - Forbid unevaluable analytical nodes such as "design a methodology / build a system / design a mechanism / produce a manual / build a framework". Every analytical node must have evidential nodes as direct or indirect upstream support.
  - **Meta nodes forbidden (hard)**: mermaid information must be concrete action+object used in solving the current query.
    - Do not write generic method nodes unrelated to content, e.g., "conduct a literature review", "deepen the research direction", "plan the next step".
    - Do not write your own Formalizer work into nodes, e.g., `identify information gaps`, `revise mermaid`, `plan next step`, `add mermaid nodes`, `generate checkpoints`.

- Grounding is the first principle: node content and logic must be supported; do not fabricate.
  - Node content may use only concepts already present and alignable in the trajectory, query, or mermaid_old; do not introduce new thematic dimensions or coined terms. Prefer existing vocabulary; do not use harder terminology for pseudo-depth.
  - Strictly forbid fabricated nodes or information: do not introduce undefined coined keywords (invented model names, abbreviations, scoring systems, method names). Widely known terms must appear verbatim in `tool_response` or mermaid_old and be locatable in the evidence fragment pointed to by `source_note`. Likewise, forbid calculation methods, numbers, formulas, or tier thresholds not validated by standard methods (i.e., obtainable via query or search).
  - Do not package for depth or professionalism: forbid named entities, abbreviations, or numeric rules absent from the evidence; forbid "terminology upgrades" (rewriting ordinary evidence as more professional absent terms, e.g., "concrete steps" -> "SOP", "compare" -> "coupling analysis", "verify truth" -> "authenticity discrimination model"). Do not import AI-coined quoted vague adjectives from trajectory answers (e.g., "loss of irony", "sense of barrenness", "mythic historicism"); use plain wording (e.g., "multi-dimensional risk assessment"); do not invent new proper nouns.

- **Edge connectivity and completeness (hard; must hold before output)**:
  - No dangling nodes: unless the graph has only one node, every node must appear in at least one edge (in-edge or out-edge).
  - No unsupported analytical nodes: every **analytical** node must have at least one in-edge from the evidential/prior analytical nodes its inference depends on; forbid analytical nodes with no upstream that connect directly to the sink.
  - No dead-end intermediates: except **evidential leaf** nodes that only collect information and are not consumed downstream, any node whose information will be used later must have at least one out-edge to a downstream consumer.
  - When revising, fix edges together with nodes: if existing nodes lack in/out edges, disagree with `source_note`, analytical nodes do not feed synthesis, or analytical nodes lack in-edges, repair the mermaid first rather than only stacking new nodes. Every newly added node must include its upstream and downstream edges at the same time.

# Self-check before output
- Recheck all node content, format, and granularity.
- Recheck the mermaid against all "Special notes".
- Check that derivation edges strictly satisfy "downstream presupposes upstream"; independently completable tasks should be parallel. Check that every analytical node has at least one in-edge.
- Check for duplicate verification / identical checkpoint conclusions for the **same fact**; merge duplicates. Do not merge distinct events, objects, or independently checkable evidence anchors merely to shorten the graph.
- Check that the rubric count matches the mermaid_new node count; revise if not.
- Check whether rubric checkpoints are overly strict beyond the query scope (e.g., requiring a specific tool or webpage); regenerate if so.
- If the mermaid uses "professional packaging" terms absent from evidence (or not widely defined), such as SOP, game theory, hedging, closed-loop mechanism, or quoted coinages, replace them with plain synonymous expressions from the evidence.

## Output format
Final output content must be in English and will be parsed as strict JSON:

```json
{
  "mermaid_new": {
    "structure": "string",
    "nodes": [
      {
        "node_id": "string",
        "node_type": "evidential | analytical",
        "check_points": ["string", "string"],
        "source_type": "tool_call_response | query_or_reason_or_answer | previous_mermaid",
        "source_note": "string"
      }
    ]
  },
  "rubric": {
    "query_id": "string",
    "criteria": [
      {
        "node_id": "A",
        "description": "string",
        "check_points": ["string"]
      }
    ]
  }
}
```
\end{lstlisting}

\subsubsection{Challenger Prompt}
\mbox{}\\[-0.6em]
\begin{lstlisting}
You are an assistant that constructs a more valuable deep research question from a mermaid and a trajectory.

You must output `query`: a more valuable deep research question. The query states what task to perform and vaguely indicates what to build on / which aspects / which methods to use.

## Known inputs
- mermaid:
{mermaid}

- trajectory:
{trajectory}


## Your workflow
First, based on the mermaid, find information in the trajectory that is related to mermaid nodes and can be deepened, then generate a new `query`. The query must be a deep research task with one shared main goal. In wording, it may contain several logically linked questions (progressive, deepening, synthetic), but must not be a set of mutually independent (or weakly related) questions. Follow the two steps below:

- Preserve difficulty while fuzzifying concrete information: the query should steer the answerer along the mermaid direction through constraints, but must not restate nodes, edges, step order, or solution modules. Omit intermediate reasoning paths and details; state only broadly which aspects to address and what problem to solve.
    - Preserve difficulty:
        - Do not write concrete results/answers already found in the trajectory as known premises; that makes the question shallower. If the trajectory already found an intermediate conclusion, keep the to-be-found object rather than reusing the conclusion. E.g., if mermaid has node A[query the strait between Madagascar and mainland Africa] and the trajectory already found it is the Mozambique Channel, do not ask "What are the spatial scale and natural characteristics of the Mozambique Channel?"; ask "What are the characteristics of the strait between Madagascar and mainland Africa?".
        - If a later mermaid node uses an earlier node's result, do not write that result as a premise; hide the later object behind the earlier node. E.g., for `A[query the 2024 Olympic men's singles table-tennis champion] --> B[query several turning points in Fan Zhendong's career]`, do not ask "What were the turning points in Fan Zhendong's career?"; ask "What were the career turning points of the 2024 Olympic men's singles table-tennis champion?".
        - Do not over-simplify; keep multi-aspect constraints. E.g., keep distinctions such as flyby / impact / soft landing / crewed landing when asking about the first lunar mission; do not collapse them into an unconstrained "what was the first mission to the Moon?".
        - Do not remove the queried object from origin nodes (out-edges only, no in-edges). E.g., for A[Identify the theme song of The Dukes of Hazzard], do not simplify to "Identify the theme song", because the search origin is The Dukes of Hazzard.
        However, for a non-origin node such as B[Search for who sings "Good Ol' Boys."] with relation A --> B, the object "Good Ol' Boys." should be hidden behind the upstream node and rewritten as "Search for who sings the theme song of The Dukes of Hazzard".
    - Omit intermediate reasoning paths and details:
        - Turn concrete information and examples into broad summaries with plain words: rewrite "Compare Luna 9 (99kg landing mass, no ascent stage), Surveyor 1 (one-way engineering verification), and Luna 15 (crashed during Apollo 11 stay) on technical parameters and mission outcomes" as "Compare prior lunar mission technology and outcomes"; rewrite long professional packaging into "analyze from recommendation mechanisms, structural privileges, and content change".
        - Rewrite toward ordinary-user language: remove redundant adjectives; drop specific terminology, formal tone, and professional methods; avoid academic phrasing such as "temporal coupling / dynamic calibration / cross-modal / semantic discrimination / tiered verification logic / final adjudication criteria"; use plain, simple, everyday wording.
        - Mermaid-invisible: the query is independent of mermaid and trajectory and is used to test whether a model can answer the task depicted by the mermaid. Therefore forbid referential phrasing such as "based on prior output / as above / as stated / previous section / previous round / final version / continue last time / according to node A / this movie".

- Guiding question: on top of fuzzifying the mermaid into a question, guide the agent toward tasks related to existing mermaid nodes (mined from the trajectory) that are still unsolved or underexplored, so as to generate a harder, concrete, reasonable, meaningful, and solvable task. Prefer small-step deepening; do not over-diverge; guide on only 1-2 aspects from the trajectory.
    - Prefer guiding questions that the trajectory already partially answers or can answer later:
        - E.g., if the trajectory only found that Zijin Mining's competitor includes Shandong Gold, do not ask "Compare Zijin Mining and Shandong Gold", but you may ask "Who are Zijin Mining's competitors?".
        - E.g., if the trajectory already retrieved revenue figures for both, you may ask "Compare Zijin Mining and Shandong Gold", even if revenue is only part of the comparison.
    - Guide in breadth and depth:
        - Breadth: if the trajectory contains information that can answer the current mermaid task more completely, include that aspect in the query.
        - Depth: if the trajectory contains information that can answer a question not yet covered by the current mermaid, and it can be merged into one larger deep research question, include that aspect.
    - Prefer "evidence-constrained deep analysis", including but not limited to:
        - Explain / verify / deepen / attribute / compare something.
        - Synthesize / reason / compare across several aspects already present in the trajectory.
    - Meaningless questions forbidden:
        - Cross-checking the same fact with multiple pieces of evidence is forbidden: "repeatedly verify / compare the same fact across multiple searches/databases and give prioritization rules / confirm via multiple aspects" is shallow and meaningless.
        - Over-autonomous questions that do not depend on evidence or reasoning are forbidden: "design methodology / design system / design mechanism / build framework / output manual" and similar abstract product-oriented tasks are meaningless. Degree questions such as "how strong / how stable is this conclusion" should also be rare unless the trajectory already provides quantified results.
        - Over-obsessing about a simple, commonly accepted fact is forbidden: do not ask why sources differ in wording, or which channel is the single most authoritative answer, unless different definitions clearly change the answer. Rewrite such questions into deeper, meaningful ones.
    - Grounding is the first principle:
        - The query may use only concepts mentioned and alignable in the trajectory and mermaid; wording may change, but do not invent new thematic dimensions, terms, named entities, abbreviations, or numeric rules absent from both. Prefer existing vocabulary. Do not ask about information completely absent from the trajectory, or unrelated to every current mermaid node.
        - When judging whether something is "involved", use what the trajectory **actually used**: concepts that appear only incidentally in a `tool_response` but were not adopted in answer/reason count as uninvolved; analytical requirements from mermaid analytical nodes must not be forced into `query` if the trajectory still lacks corresponding evidence.
        - Strictly forbid fabricated information: do not introduce undefined coined keywords. Widely known terms must come from `tool_response` verbatim or from mermaid node descriptions. Forbid calculation methods, numbers, formulas, or tier thresholds not validated by standard methods.
        - Do not package for depth or professionalism:
            - No "terminology upgrades" (e.g., "concrete steps" -> "SOP", "compare" -> "coupling analysis", "verify truth" -> "authenticity discrimination model").
            - Do not import AI-coined quoted vague adjectives from trajectory answers; use plain wording; do not invent new proper nouns.

**Special note:** If the trajectory has no further mineable information, you **must** only fuzzify and not add a guiding question. If the trajectory is mainly fact-list retrieval, the question is already deep and meaningful, or further guiding would become multiple parallel independent questions, you **must** also only fuzzify. Prefer a thorough look-up/clarification question over an analytical `query` unsupported by the trajectory.

## Output format (strict JSON):

```json
{
  "query": "string"
}
```

## Self-check before output
- If the query contains "based on prior output / as above / previous section / previous round / continue last time / final version", rewrite as a self-contained version.
- If the query is same-fact verification / multi-source wording disputes / over-autonomous product design, rewrite into a meaningful question.
- If the query writes trajectory-found concrete results as known premises, rewrite to keep the to-be-found object and hide that result.
- If the query is multiple small parallel subquestions without one shared main goal, rewrite into one deep research question with a shared goal.
- Check every analysis dimension/constraint in `query`: if it cannot be found in the trajectory's `tool_response` or answer/reason, delete it; if only fact-clarification remains, switch to fuzzify-only without guiding analysis.
- If `query` asks for synthesis / significance / pattern induction but the trajectory only enumerates facts without synthesis, rewrite into a clarification question covering only the retrieved facts.
\end{lstlisting}

\subsubsection{Without-Hints Query Prompt}
\mbox{}\\[-0.6em]
\begin{lstlisting}
You are an assistant that rewrites a deep research question into a shorter, more colloquial user-facing question.

Given a `query`, output `query_fuzzified` that preserves the same goal, semantic alignment, and all subquestions. `query_fuzzified` should only ask the question itself, without constraints or solution pathways.

## Input query
{query}

## Rewrite rules
- **Semantic alignment, no dropped questions**:
    - Do not omit the core queried object: e.g., "How is the protagonist's character shaped in The Count of Monte Cristo?" must not become "How is the protagonist's character shaped?", which becomes vague.
    - Do not change the main object of the query: if the query asks how to understand a show's most famous theme song against a broader music background and creative/performing roles, rewrite to "How should we understand the show's most famous theme song?", rather than elevating the background into a co-equal asked object or retaining scaffolding such as "considering creative/performing roles and broader film/TV music history".
    - Do not drop subquestions: a long query that asks both for intrinsic causes and clinical implications may be shortened, but must keep both subgoals rather than keeping only the clinical part.
- **Question simplification**:
    - Width fuzzification: remove professional terms, methods, and execution details. E.g., "Why is one-to-one assessment needed for air-combat pilots' flying skills?" -> "Why do air-combat pilots' flying skills need to be assessed?"
    - Process fuzzification: remove pathways. E.g., "How to shape the protagonist of The Count of Monte Cristo through story, language, and behavior?" -> "How is the protagonist's character shaped in The Count of Monte Cristo?"
- **Width fuzzification (most important), remove constraints**:
    - "Do not drop subquestions" means preserve semantic coverage of every goal in the query -- **not** copy the query's parallel lists verbatim into `query_fuzzified`.
    - Delete facet/dimension constraints: remove A, B, C from "from aspects A, B, C" and keep only the core task. E.g., "analyze xxx from algorithm mechanisms, special identity, and content change" -> "analyze the cause of xxx"; "analyze under flyby / impact / soft landing / crewed landing definitions..." -> "analyze under different definitions...".
    - **No serial listing (including pairs)**: do NOT use commas / semicolons / `and` / `or` / `as well as` / `including` / `such as` / `both ... and` to list **2+ parallel** analysis dimensions, plotlines, mechanisms, levels, conditions, or channels. Even "A and B" or "A, B, or C" must become one umbrella phrase + task verb (e.g., "what mainly drives X" / "how Y is handled").
    - **Delete scaffold and background constraints**: drop `based on`, `depending on`, `in terms of`, `at the legal and practical level`, `from ... and ... aspects`, `considering ... and ...`, `in the context of`, `against the backdrop of`, `broader ... history`, etc. when they only enumerate angles, evidence, or paired "specific sub-factors + broader history/context". Keep only the core ask; background stays implicit.
    - **Delete decorative modifiers**: drop style/evaluative adjectives (freer, character-driven, shifting tone, nuanced, multifaceted) when the question stays complete without them.
- **Process fuzzification**:
    - Delete process constraints, step hints, and solution pathways: `Focus on`, `Analyze by`, `Trace`, `Use evidence`, `Research`, `Investigate`, `when you separate`, and `Please compare` must not appear.
    - Delete scoring conditions, source requirements, and extra boundary constraints; do not add new constraints absent from the query.
    - Hide solution order: if the query has layered questions, embed dependent subquestions into one main question without exposing "first A, then B, then C". E.g., "Who has the most followers on platform X in 2026? Why the gap?" -> "Why does the most-followed account on platform X in 2026 lead other top accounts by such a wide margin?"
- **Expression form**:
    - Complete, natural user question; prefer **1 sentence**, at most 2, never more than 3.
    - Keep the core asked content; summarize other proper names when possible.
    - Prefer one main question that implicitly contains all subquestions; colloquial, plain, low jargon.

## Output format (strict JSON)
```json
{
  "query_fuzzified": "string"
}
```

## Self-check before output (hard gates -- rewrite if any fail)

1. **Main object**: any confusion about the query's main object? any dropped subquestion?
2. **Comma segments**: split on `,`; count **<= 3** (strongly prefer **<= 2**)? If it still lists parallel items, use umbrella phrasing.
3. **Checklist smell**: still 2+ parallel facets (`A and B`, `A, B, or C`)? `based on` / `in terms of` / `at ... levels` / `including` / `considering ... and ... broader ... history` / `roles behind`? stacked style adjectives? Delete or merge all.
4. **Sentences & question marks**: split on `.!?` + space; **<= 3 sentences**? **<= 2 `?`**? Merge to one sentence when possible.
5. **Length ratio**: `len(query_fuzzified) / len(query) <= 0.70`? If not, keep deleting modifiers, lists, and method words.
6. **Process leakage**: can a reader infer step order? Any First/Then/Investigate/Analyze by? Rewrite.
7. **Read-aloud test**: if it sounds like a research rubric or outline rather than a real-user question, rewrite.
\end{lstlisting}

\subsubsection{Assigned-Topic Query Prompt}
\mbox{}\\[-0.6em]
\begin{lstlisting}
The mermaid below describes my completed deep research content and conclusions (nodes capture the research process and information; check_points are the key findings/evidence).

## My research (mermaid)
{mermaid}

What is the most plain, fitting research title `title_query` for this work? `title_query` must be a **one-sentence informational title** for a report or research project, such that the mermaid covers the title's research scope and does not go beyond it. Not a question, outline, or checklist.

## Rules
- **Title form, not a question**: noun phrase or short declarative heading; do not start with Why / How / What / Which; do not end with "?".
- **Keep a searchable research subject (most important)**:
  - From the mermaid and check_points, identify the **concrete main object of study** (work title, season/episode, full person name, historical event, place + time, etc.) and include it in the title.
  - **Do not** use nicknames or subplot lines unknown to the general public as the research subject.
- **Informational and non-literary**:
  - Use **neutral, plain, verifiable** informational words; **do not** use literary, rhetorical, or emotional wording.
  - **Do not** rewrite check_points into prettier synonyms; prefer plain nouns already in the mermaid / check_points.
  - It should read like a **research project title**, not a review headline, essay title, or marketing copy.
- **Let analytical conclusions set the tone**: while keeping the subject, summarize the most downstream analytical conclusion -- the research purpose of the whole mermaid; do not write an evidential retrieval checklist.
- **Shorter than the research content**: merge dimensions; do not enumerate item by item; usually 10-18 English words (slightly longer when the subject string is included).
- **No solution pathways and no mermaid internals**: no "first A then B", node IDs, or retrieval steps. Do not write node A/B, and do not use pronouns such as this/that.

## Self-check before output
- After reading the title, can someone tell **which work / episode / person / event** is studied? If not, add the subject.
- Any metaphor, hype adjective, or emotional judgment? Replace with neutral informational wording.
- Is it one highly condensed sentence that summarizes the mermaid's research content?

## Output format (strict JSON)
```json
{
  "title_query": "string"
}
```
\end{lstlisting}

\subsection{Hyperparameters Settings}

NQ-Open seed sampling uses seed $42$ to sample $1{,}000$ seed queries.
During construction, GPT-5.5 serves as the Explorer, Formalizer, and Challenger.
The reviser uses temperature $0.8$ and at most $100{,}000$ output tokens.
Each case allows at most $100$ evolution rounds and at most $200$ Explorer turns
per round.
For agent-mode solving, the solver temperature and thinking settings follow the
API defaults, with at most $65{,}536$ output tokens and at most $200$ turns.
For evaluation, the judge is Qwen3.7-Max and receives only the final response
and the rubrics.

\subsection{Computing Environment}

The search environment provides two tools, \texttt{google\_search} and
\texttt{web\_scrape}.
The former calls Google Search via the Serper API and returns retrieval results,
knowledge-graph items, and related questions.
The latter preferentially uses Jina AI to fetch and parse webpage body text.
Compute cost of construction is dominated by API calls.
The full construction run took about 168 hours and cost about \$10{,}000.
Orchestration and experimental runs used a MacBook Air (Apple M4, 10 cores:
4 performance + 6 efficiency, integrated GPU), 16~GB memory, and macOS~15.7.1
(arm64).

\section{Human Review Criteria}
\label{app:review_criteria}

The main paper reports the review setup, consistency, and aggregate outcomes.
Table~\ref{tab:review_criteria} lists the eleven review questions and the
poor/medium/good criteria used by the three researchers.

\begin{table*}[t]
\centering
\footnotesize
\begin{tabular}{p{0.24\textwidth}p{0.24\textwidth}p{0.24\textwidth}p{0.24\textwidth}}
\toprule
Question & Poor & Medium & Good \\
\midrule
Is $q_T^{\mathrm{h}}$ meaningful? & Invalid or lacks research value & Meaningful but mostly factoid-like & Meaningful research question \\
Is $q_T^{\mathrm{h}}$ hard enough? & Answerable by simple lookup & Some difficulty but limited reasoning & Requires evidence collection and rigorous analysis \\
Does $q_T^{\mathrm{o}}$ keep the original meaning? & Changes topic or loses key subquestions & Keeps main goal with minor drift & Preserves the intent of $q_T^{\mathrm{h}}$ \\
Is $q_T^{\mathrm{o}}$ concise enough? & Too long or detail constrained & Shorter but still gives hints & Short and natural with a hidden solution path \\
Does $q_T^{\mathrm{a}}$ match the main topic? & Off target or adds irrelevant content & Related but imprecise & Captures the main research focus \\
Is $G_T$ correct? & Invalid structure or incorrect node content & Mostly correct with some flaws & Correct relationships and node content \\
Is $R_T$ discriminative? & Vague and hard to score & Partly clear with ambiguity & Clearly separates high and low scores \\
Does $R_T$ cover the important points? & Misses important aspects & Covers main aspects with gaps & Covers major required aspects \\
Is the strictness of $R_T$ appropriate? & Out of scope or overly strict & Mostly appropriate with edge cases & Neither out of scope nor overly strict \\
Are the weights in $R_T$ reasonable? & Weights do not match importance & Has minor weighting imbalance & Weights reflect importance \\
Are the checkpoints in $R_T$ verifiable? & Has unverifiable checkpoints & Mostly checkable with unclear wording & All checkpoints are verifiable \\
\bottomrule
\end{tabular}
\caption{Human quality-review questions and scoring criteria.}
\label{tab:review_criteria}
\end{table*}

\section{Failed Evolution Cases}
\label{app:failed_evolution}

After evolution, 963 of the $1{,}000$ seed queries converged successfully and
37 failed.
Besides successful convergence under the stopping criterion in the main paper,
task evolution may fail in the following cases.
\begin{itemize}
    \item \textbf{Invalid DAG structure.}
    We parse the constructed DAG and compute its depth and node count.
    If either value cannot be extracted, the case is treated as a failed
    evolution.
    \item \textbf{Overly long single reasoning chain.}
    We identify the longest contiguous chain of analytical nodes.
    If it contains more than five nodes, we stop evolution because the process
    is extending one reasoning chain rather than constructing balanced
    multi-source research.
    \item \textbf{Pipeline or tool failure.}
    A case fails when the LLM cannot return a valid reply because of
    malformed inputs, network issues, or related runtime faults.
    \item \textbf{No convergence within the round budget.}
    A case fails when it reaches the maximum of 100 rounds without satisfying
    the DAG-stability stopping criterion.
\end{itemize}

\section{Case Analyses}
\label{app:case_analyses}

\subsection{Strengths}

\textbf{Natural query wording.}
After evolution from ``\emph{kutchhi language is a mix of which two languages}'',
the round-9 query without hints asks how Kutchi actually stands in relation to
Sindhi and Gujarati, and why some of its varieties sound closer to one than to
the other.
It reads like a genuine question from a curious reader following a line of
thought, rather than a rigid template that merely stacks retrieval steps.

\textbf{Tight subquestion logic.}
From simple query ``\emph{what is the function of a growth plate}'', evolution
reaches a clinical research question that runs from growth-plate structure and
endochondral ossification, to why pediatric plates are mechanically vulnerable,
to how trauma disrupts them, and only then to divergent long-term outcomes and
the choice of imaging, treatment, and follow-up.
The subquestions are not loosely connected but chained by the DAG structure,
where each step is a genuine prerequisite for the next.
Answering any late node therefore requires the reasoning accumulated in its
ancestors, so the subquestions form one coherent line of inquiry that cannot be
shortcut by retrieving isolated facts.

\textbf{High difficulty and professionalism.}
From simple query ``\emph{when did residents of puerto rico became us citizens}'',
whose answer is a single date, the final task instead traces how Puerto Rican
U.S.\ citizenship evolved from collective territorial status to jus soli
birth-based acquisition.
Solving it requires genuine legal expertise, not a single lookup.
One must read and reconcile the Foraker Act, the Jones Act, and the 1934--1952
legislative history, and distinguish statutory grants of citizenship from
constitutional guarantees with the precision expected of a domain specialist.
Such depth of investigation across primary legal sources is what makes the task
hard for models to answer.

\textbf{High verifiability without format rigidity.}
From simple query ``\emph{what us ships were sunk in pearl harbor}'', the final
task separates physical damage on attack day from postwar salvage, repair, and
re-hulling and reclassifies capital ships and destroyers accordingly.
In particular, we impose no requirement on writing style or answer format, and
only check whether the model response covers the checkpoints that genuinely
contribute to resolving the task.
Rubrics thus specify the verifiable information the task needs rather than
prescribing an answer template.

\begin{figure}[t]
\centering
\includegraphics[width=\linewidth]{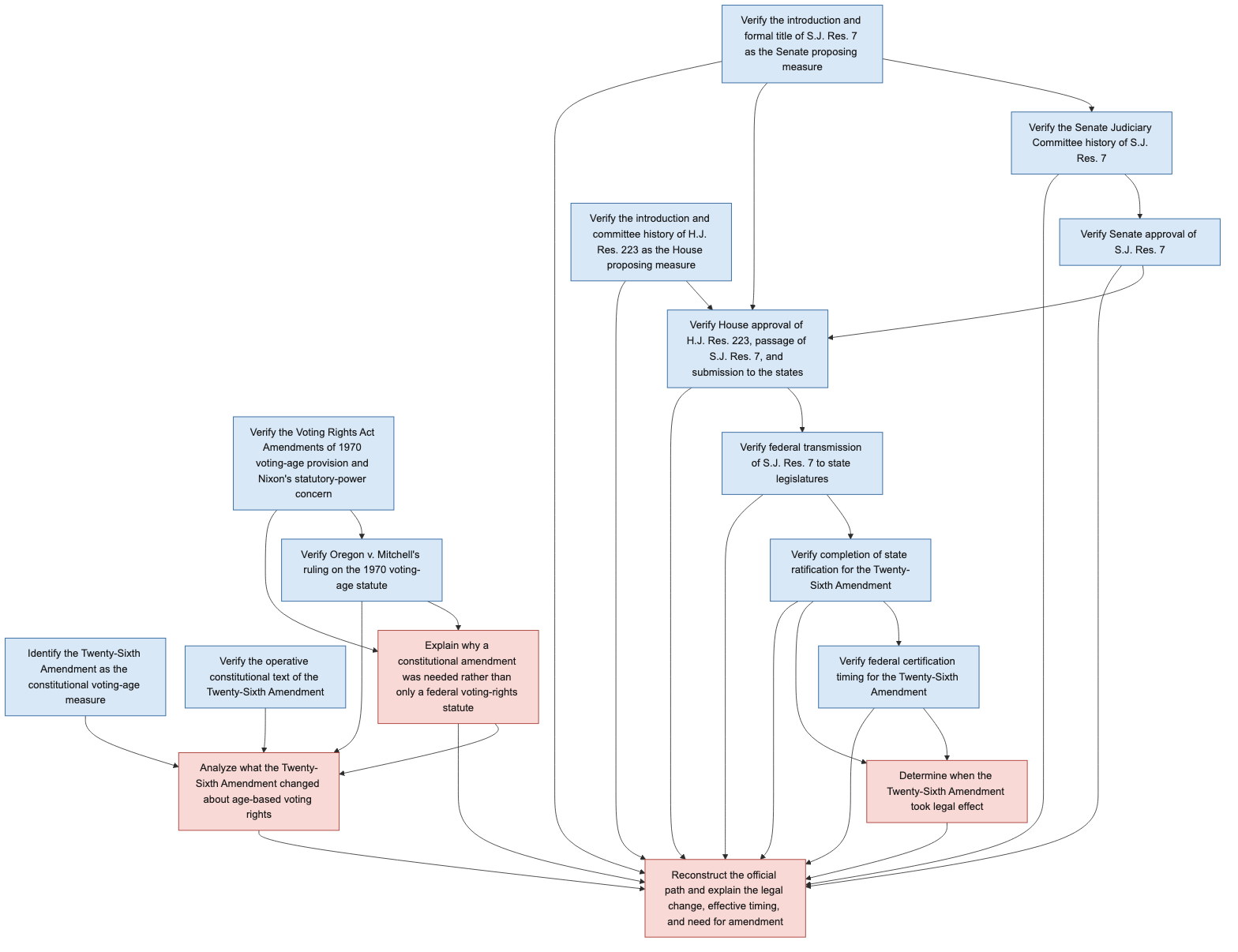}
\caption{Final task DAG for \texttt{nq\_open\_seed\_0395}.}
\label{fig:appendix_dag_0395}
\end{figure}

\subsection{Weaknesses}

The current benchmark score measures coverage and correctness of task-specific
evidential and analytical checkpoints.
It does not separately measure citation count, citation rate, citation format,
matching to a prescribed source list, prose style, document layout, or
tool-call efficiency.
These dimensions can affect the quality of a real deep research report, but
they are outside the current rubric score.

\subsection{An Example of Task Evolution}
\label{app:artifact_example}

We present the successfully converged case
\texttt{nq\_open\_seed\_0395}.
Its final DAG has depth 9 and 16 nodes, and its stopping reason is that the
complete DAG remained unchanged for two consecutive rounds.

\textbf{Simple query $q_1$.}
\emph{which legislation was responsible for changing the voting age in the
united states brainly}

\textbf{Query with hints $q_T^{\mathrm{h}}$.}
Research the constitutional change that lowered the U.S.\ voting age to 18 and
reconstruct its official legal path.
Identify the amendment, explain what it changed in voting rights, trace how the
relevant congressional proposal moved through Congress and the states, and
determine when it became legally effective compared with when it was formally
certified.
Also analyze why an ordinary federal voting-rights statute was not enough,
using the earlier statutory attempt, the Supreme Court's treatment of it, and
the resulting practical and constitutional problems as evidence.

\textbf{Query without hints $q_T^{\mathrm{o}}$.}
How did the U.S.\ constitutionally lower the voting age to 18, when did it take
legal effect compared with formal certification, and why wasn't a regular
federal voting-rights law enough?

\textbf{Assigned-topic query $q_T^{\mathrm{a}}$.}
S.J.\ Res.\ 7 and the Twenty-Sixth Amendment's Path, Effective Date, and
Voting-Age Effect

\textbf{Final DAG.}
The graph links the 1970 statutory attempt and \emph{Oregon v.\ Mitchell} to
the need for a constitutional amendment, while separately tracing S.J.\ Res.\ 7
and H.J.\ Res.\ 223 through Congress, state ratification, and federal
certification.
These branches feed analyses of the amendment's legal effect and final
chronological synthesis.
Figure~\ref{fig:appendix_dag_0395} shows the final DAG (node text in the original
English, blue for evidential nodes and red for analytical nodes, with node IDs
prefixed in each label).

Below, Table~\ref{tab:appendix_rubric} gives the complete final rubric.
Weights are shown to two decimal places and sum to one. Table~\ref{tab:appendix_trace} summarizes the scale and evolution focus of
$G_t$ across rounds.

{\footnotesize
\setlength{\LTpre}{\textfloatsep}
\setlength{\LTpost}{\textfloatsep}
\begin{longtable}{@{}cp{0.23\textwidth}p{0.58\textwidth}r@{}}
\toprule
Node & Description & Checkpoints & Weight \\
\midrule
\endfirsthead
\multicolumn{4}{c}{\tablename\ \thetable\ (continued)} \\
\toprule
Node & Description & Checkpoints & Weight \\
\midrule
\endhead
\bottomrule
\endfoot
\bottomrule
\caption{Final rubric set $R_T$ for \texttt{nq\_open\_seed\_0395}.}
\label{tab:appendix_rubric} \\
\endlastfoot
A & Identify the Twenty-Sixth Amendment as the constitutional voting-age measure &
(1) Identifies the Twenty-Sixth Amendment as the constitutional measure that lowered the voting age to 18.
(2) Does not confuse it with the Voting Rights Act Amendments of 1970.
(3) Gives March 23 and July 1, 1971 as congressional passage and ratification dates. &
$0.01$ \\
\addlinespace
G & Verify the amendment's operative constitutional text &
(1) States the substance of Section 1 and its application to both the United States and every State.
(2) States Congress's Section 2 enforcement power.
(3) Connects the text to citizens aged 18 or older. &
$0.01$ \\
\addlinespace
K & Verify the 1970 statutory attempt and Nixon's concern &
(1) Explains that the 1970 Act attempted to set age 18 for federal, state, and local elections.
(2) Identifies it as an ordinary federal statute.
(3) States that Nixon signed it on June 22, 1970 while questioning Congress's statutory authority and favoring an amendment. &
$0.03$ \\
\addlinespace
L & Verify \emph{Oregon v.\ Mitchell} &
(1) Identifies 400 U.S.\ 112 (1970).
(2) States that Congress could set age 18 for federal elections but not state and local elections by statute.
(3) Explains why the decision prevented one uniform statutory voting age. &
$0.02$ \\
\addlinespace
M & Explain why a constitutional amendment was needed &
(1) Connects the constitutional need to the statutory limit after \emph{Oregon v.\ Mitchell}.
(2) Describes the dual-age voting problem and administrative burden.
(3) Gives at least one practical problem, such as confusion, delay, fraud risk, impracticality, or expense. &
$0.03$ \\
\addlinespace
B & Verify S.J.\ Res.\ 7's introduction and title &
(1) Identifies S.J.\ Res.\ 7 as the Senate proposing measure.
(2) States that Jennings Randolph introduced it on January 25, 1971 in the 92nd Congress.
(3) Gives the formal title or its substance and the three-fourths ratification requirement. &
$0.14$ \\
\addlinespace
O & Verify the Senate Judiciary Committee history &
(1) States that the committee favorably reported S.J.\ Res.\ 7 in March 1971.
(2) Identifies S.\ Rep.\ No.\ 92-26 when report numbers are discussed.
(3) Connects the committee rationale to age discrimination and the post-\emph{Mitchell} administrative problem. &
$0.08$ \\
\addlinespace
C & Verify Senate approval &
(1) States that the Senate passed S.J.\ Res.\ 7 on March 10, 1971.
(2) Gives the unanimous 94--0 vote.
(3) Preserves the correct order before House submission to the states. &
$0.06$ \\
\addlinespace
N & Verify H.J.\ Res.\ 223's House history &
(1) Identifies the nearly identical House companion.
(2) States that Emanuel Celler introduced it on January 29, 1971.
(3) Gives its House Judiciary Committee treatment and explains that the House ultimately passed S.J.\ Res.\ 7. &
$0.06$ \\
\addlinespace
D & Verify House passage and state submission &
(1) States that the House approved H.J.\ Res.\ 223 on March 23, 1971 by 401--19.
(2) States that the House then passed S.J.\ Res.\ 7.
(3) Identifies March 23 as the date of submission to the states. &
$0.11$ \\
\addlinespace
P & Verify federal transmission to state legislatures &
(1) Explains transmission after final congressional passage.
(2) States that GSA Administrator Robert L.\ Kunzig sent certified copies to governors on March 24, 1971.
(3) Accurately describes the governors' and federal certifying official's roles. &
$0.09$ \\
\addlinespace
E & Verify completion of state ratification &
(1) States that the three-fourths threshold was reached on July 1, 1971.
(2) States that 38 of 50 states were required and identifies North Carolina as the 38th state when discussed.
(3) Distinguishes completed ratification from the later ceremony. &
$0.07$ \\
\addlinespace
F & Verify federal certification timing &
(1) States that Robert L.\ Kunzig certified the amendment on July 5, 1971.
(2) Identifies 85 Stat.\ 829--830 when publication details are included.
(3) Explains that Nixon's ceremonial signature was not legally necessary. &
$0.03$ \\
\addlinespace
H & Analyze the age-based voting-rights change &
(1) Explains the constitutional prohibition on age-based denial or abridgment for citizens 18 or older.
(2) Applies the rule to federal and state governments.
(3) Connects the change to a national voting age of 18 and elimination of the dual-age system. &
$0.02$ \\
\addlinespace
I & Determine the legal effective date &
(1) States that the amendment became legally effective on July 1, 1971.
(2) Distinguishes the July 5 certification date.
(3) Explains why ratification, rather than ceremonial signatures, controls under Article V. &
$0.02$ \\
\addlinespace
J & Reconstruct the complete official path and legal effect &
(1) Gives the chronology from the two proposing measures through committee action, congressional passage, transmission, ratification, and certification.
(2) Includes the key dates January 25, January 29, March 10, March 23, July 1, and July 5, 1971.
(3) Explains the amendment text, effective-date distinction, statutory insufficiency, \emph{Oregon v.\ Mitchell}, and the dual-age problem without treating the statute or certification ceremony as the amendment or ratification event. &
$0.22$ \\
\end{longtable}}

{\footnotesize
\setlength{\LTpre}{\textfloatsep}
\setlength{\LTpost}{\textfloatsep}
\begin{longtable}{@{}crrp{0.68\textwidth}@{}}
\toprule
Round $t$ & Depth & Nodes & Main evolution focus \\
\midrule
\endfirsthead
\multicolumn{4}{c}{\tablename\ \thetable\ (continued)} \\
\toprule
Round $t$ & Depth & Nodes & Main evolution focus \\
\midrule
\endhead
\bottomrule
\endfoot
\bottomrule
\caption{Evolution trace for \texttt{nq\_open\_seed\_0395}.}
\label{tab:appendix_trace} \\
\endlastfoot
1 & 2 & 4 & Identify the Twenty-Sixth Amendment, its text, and basic passage and ratification dates. \\
2 & 3 & 5 & Reconstruct the formal proposing measure and distinguish ratification from certification. \\
3 & 7 & 10 & Add S.J.\ Res.\ 7, the statutory background, \emph{Oregon v.\ Mitchell}, and the need for an amendment. \\
4 & 7 & 13 & Add House and Senate paths, state submission, and the dual-age voting problem. \\
5 & 7 & 13 & Strengthen primary-source requirements while retaining the same graph scale. \\
6 & 7 & 14 & Add federal transmission and connect statutory limits to a uniform nationwide rule. \\
7 & 8 & 15 & Deepen the official congressional, ratification, and certification chronology. \\
8 & 9 & 16 & Complete the two proposing-measure branches and legal-effect analysis. \\
9 & 9 & 16 & Preserve the complete graph while simplifying the final query wording. \\
10 & 9 & 16 & Confirm graph stability and retain the final DAG, rubric set, and assigned-topic query. \\
\end{longtable}}

\subsubsection{Queries with Hints, Without-Hints Queries, and Assigned-Topic Queries Across Rounds}

\paragraph{Round $t=1$.}
Identify the Twenty-Sixth Amendment, its text, and basic passage and
ratification dates.

$q_{1}$: which legislation was responsible for changing the voting age in the united states brainly

$q_{1}^{\mathrm{o}}$: \na

$q_{1}^{\mathrm{a}}$: \na

\paragraph{Round $t=2$.}
Reconstruct the formal proposing measure and distinguish ratification from
certification.

$q_{2}$: Determine which U.S. constitutional or legislative measure changed the voting age in the United States, and support the answer by checking authoritative information about its official title, when it was approved and made effective, and what its text says about voting rights based on age.

$q_{2}^{\mathrm{o}}$: Which U.S. measure changed the voting age, and when and how did it officially change age-based voting rights?

$q_{2}^{\mathrm{a}}$: The 1971 Twenty-Sixth Amendment and the U.S. Voting Age Change to Eighteen

\paragraph{Round $t=3$.}
Add S.J.\ Res.\ 7, the statutory background, \emph{Oregon v.\ Mitchell}, and the
need for an amendment.

$q_{3}$: For the U.S. voting age change, determine which constitutional measure made the change and reconstruct its official path from congressional proposal to becoming part of the Constitution. Use authoritative sources to identify the formal proposing measure, key congressional approval dates, ratification and certification timing, and the operative constitutional text, and explain how these pieces of evidence show both what changed and when the change took legal effect.

$q_{3}^{\mathrm{o}}$: Which constitutional amendment lowered the U.S. voting age, how did it become part of the Constitution, and when did it legally take effect?

$q_{3}^{\mathrm{a}}$: S.J. Res. 7 and the Twenty-Sixth Amendment's Voting-Age Change and Effective Date

\paragraph{Round $t=4$.}
Add House and Senate paths, state submission, and the dual-age voting problem.

$q_{4}$: Research the official legal history of the Twenty-Sixth Amendment as the U.S. constitutional measure on voting age. Determine how the proposing measure, including S.J. Res. 7 and its formal title, moved through Congress and the states, when the amendment became legally effective versus when it was federally certified, and what its operative text changed about age-based voting rights. Also explain, using the surrounding statutory and court background discussed in official sources, why a constitutional amendment was needed rather than only a federal voting-rights statute.

$q_{4}^{\mathrm{o}}$: When and how did the Twenty-Sixth Amendment become part of U.S. law, what did it change about age-based voting rights, and why was a constitutional amendment needed instead of just a federal law?

$q_{4}^{\mathrm{a}}$: The Twenty-Sixth Amendment's Adoption, Voting-Age Effect, and Need After Oregon v. Mitchell

\paragraph{Round $t=5$.}
Strengthen primary-source requirements while retaining the same graph scale.

$q_{5}$: Using official constitutional, congressional, judicial, and archival sources, reconstruct how the United States made voting rights for citizens eighteen or older constitutionally protected. Identify the relevant amendment and the congressional measure that proposed it, trace its path through Congress and state ratification, determine the difference between the amendment's legal effectiveness and its later federal certification, and explain why an ordinary federal voting-rights statute was not enough after the earlier statutory attempt and Supreme Court review.

$q_{5}^{\mathrm{o}}$: How did the U.S. make voting rights for citizens 18 and older constitutionally protected, when did that protection legally take effect versus get certified, and why wasn't a regular federal voting-rights law enough?

$q_{5}^{\mathrm{a}}$: The Twenty-Sixth Amendment's adoption, voting-age effect, and need after Oregon v. Mitchell

\paragraph{Round $t=6$.}
Add federal transmission and connect statutory limits to a uniform nationwide
rule.

$q_{6}$: Investigate how the United States legally lowered the voting age to eighteen through constitutional change rather than relying only on ordinary federal legislation. Identify the relevant constitutional amendment and proposing measures, trace its path through Congress and state ratification, determine when it became legally effective versus when it was federally certified, and explain why the earlier statutory attempt and the Supreme Court litigation over it made a constitutional amendment necessary. Use official constitutional text, Statutes at Large materials, congressional history, Supreme Court sources, and ratification or certification records to support the reconstruction.

$q_{6}^{\mathrm{o}}$: How did the U.S. lower the voting age to 18 through a constitutional amendment, when did it legally take effect versus get certified, and why wasn't the earlier federal law enough?

$q_{6}^{\mathrm{a}}$: The Twenty-Sixth Amendment's official path, legal effect, and need after Oregon v. Mitchell

\paragraph{Round $t=7$.}
Deepen the official congressional, ratification, and certification chronology.

$q_{7}$: Research how the United States legally lowered the voting age to 18 through the constitutional amendment process. Identify the relevant constitutional amendment and reconstruct its official path from the earlier federal statutory attempt through Supreme Court review, congressional proposal, state ratification, and federal certification. Explain what legal change the amendment made, when it took effect as opposed to when it was formally certified, and why ordinary voting-rights legislation was not enough to create a uniform rule for federal, state, and local elections. Use official constitutional text, Statutes at Large materials, congressional records or committee materials, the relevant Supreme Court decision, and federal certification records where appropriate.

$q_{7}^{\mathrm{o}}$: How did the U.S. lower the voting age to 18 through a constitutional amendment, when did the change take effect, and why couldn't ordinary voting-rights laws create one nationwide rule?

$q_{7}^{\mathrm{a}}$: The Twenty-Sixth Amendment's Official Path, Voting-Age Legal Effect, and Need After Oregon v. Mitchell

\paragraph{Round $t=8$.}
Complete the two proposing-measure branches and legal-effect analysis.

$q_{8}$: Reconstruct the official legal and legislative path by which the United States lowered the voting age through the constitutional amendment process. Identify the relevant constitutional measure and its operative text, trace how the congressional proposal moved through the Senate and House and then to state ratification and federal certification, determine when the amendment became legally effective, and explain why a constitutional amendment was needed after Congress had already attempted to lower the voting age by statute. Use primary legal materials where possible, including the statutory voting-age provision, the Supreme Court's ruling on it, congressional reports or records, Statutes at Large materials, and the federal certification record.

$q_{8}^{\mathrm{o}}$: How did the Twenty-Sixth Amendment legally lower the U.S. voting age, when did it take effect, and why was it needed after Congress had already tried to do the same by statute?

$q_{8}^{\mathrm{a}}$: S.J. Res. 7 and the Twenty-Sixth Amendment's ratification, legal effect, and need beyond the 1970 Voting Rights Act

\paragraph{Round $t=9$.}
Preserve the complete graph while simplifying the final query wording.

$q_{9}$: Research the official legal path by which the United States extended voting rights to citizens aged eighteen or older nationwide. Identify the constitutional amendment and the congressional measures that proposed it, trace the main Senate and House actions through submission to the states, follow the state ratification and federal certification process, and determine when the change legally took effect. Also explain why an ordinary federal voting-rights statute was not enough, using the earlier voting-age statute, the Supreme Court's ruling on it, and the practical problem of different voting ages for federal versus state and local elections as the main points of analysis.

$q_{9}^{\mathrm{o}}$: How did the U.S. legally lower the nationwide voting age to 18, when did it take effect, and why did it require a constitutional amendment rather than just a federal law?

$q_{9}^{\mathrm{a}}$: The Twenty-Sixth Amendment's Official Path, Voting-Age Change, and Effective Date After Oregon v. Mitchell

\paragraph{Round $t=10$.}
Confirm graph stability and retain the final DAG, rubric set, and assigned-topic
query.

$q_{10}$: Research the constitutional change that lowered the U.S. voting age to 18 and reconstruct its official legal path. Identify the amendment, explain what it changed in voting rights, trace how the relevant congressional proposal moved through Congress and the states, and determine when it became legally effective compared with when it was formally certified. Also analyze why an ordinary federal voting-rights statute was not enough, using the earlier statutory attempt, the Supreme Court's treatment of it, and the resulting practical and constitutional problems as evidence.

$q_{10}^{\mathrm{o}}$: How did the U.S. constitutionally lower the voting age to 18, when did it take legal effect compared with formal certification, and why wasn't a regular federal voting-rights law enough?

$q_{10}^{\mathrm{a}}$: S.J. Res. 7 and the Twenty-Sixth Amendment's Path, Effective Date, and Voting-Age Effect

\end{document}